\documentclass[conf]{new-aiaa}

\usepackage[utf8]{inputenc} % allow utf-8 input
\usepackage[T1]{fontenc}    % use 8-bit T1 fonts
\usepackage{inconsolata}    % improve the aesthetics of text in the typewriter font.
\usepackage{hyperref}       % hyperlinks
\usepackage{url}            % simple URL typesetting
\usepackage{booktabs}       % professional-quality tables
\usepackage[version=4]{mhchem}
\usepackage{siunitx}
\usepackage{tabularx}
\usepackage{nicefrac}       % compact symbols for 1/2, etc.
\usepackage{microtype}      % microtypography
\usepackage{graphicx}
\usepackage{subfigure}
\usepackage{wrapfig}
\usepackage{eso-pic}         % for icon on the first page
\usepackage{array}           % for table of half page
\usepackage{caption}
\usepackage{lipsum}
\usepackage{multirow}
\usepackage[listings,skins,breakable]{tcolorbox}
\usepackage{amsmath}

\usepackage{amssymb}
\usepackage{mathtools}
\usepackage{amsthm}
\usepackage{algorithm}      % algorithms
\usepackage{algorithmic}    % algorithms
\usepackage[table,xcdraw,svgnames]{xcolor}
\usepackage{bbding}         % for checkmark ans xmark
\usepackage{duckuments}     % for example-image-duck/a
\usepackage{bm}
\usepackage{CJKutf8} % for kerean characters
\usepackage{authblk}

\usepackage[capitalize]{cleveref}
\theoremstyle{plain}

\theoremstyle{definition}

\theoremstyle{remark}

\newcommand*{\rom}[1]{\expandafter\@slowromancap\romannumeral #1@}

\title{Position: Life-Logging Video Streams Make the Privacy-Utility Trade-off Inevitable}

\author[1,2]{\textbf{Tianyuan Zou}\footnote{Work done during internship at RayNeo.AI.}}
\author[2]{\textbf{Liang Yue}}
\author[1,3]{\textbf{Yang Liu}}
\author[1]{\textbf{Ya-Qin Zhang}} 
\author[1,2,3]{\textbf{Sijie Cheng}\footnote{Corresponding author <\textit{chengsj@rayneo.com}>.}}

\affil[1]{\footnotesize Institute for AI Industry Research, Tsinghua University, Beijing, China}
\affil[2]{\footnotesize RayNeo.AI, Shenzhen, China}
\affil[3]{\footnotesize Department of Computer Science and Technology, Tsinghua University, Beijing, China}

\begin{document}
\AffiliationIcon

\maketitle

\begin{abstract}
  % The abstract paragraph should be indented \nicefrac{1}{2}~inch (3~picas) on both theceft- and right-hand margins. Use 10~point type, with a vertical spacing (leading) of 11~points. The word \textbf{Abstract} must be centered, bold, and in point size 12. Two line spaces precede the abstract. The abstract must be limited to one paragraph.
  With the growing prevalence of always-on hardware such as smart glasses, body cameras, and home security systems, life-logging visual sensing is becoming inevitable, forming the backbone of persistent, always-on AI systems. Meanwhile, recent advances in proactive agents and world models signal a fundamental shift from episodic, prompt-driven tools to next-generation AI systems that continuously perceive and react to the physical world. 
  % Although life-logging video streams can substantially improve utility, they also introduce unprecedented privacy risks by revealing geographical routines, behavioral patterns, emotional states, social interactions, and other sensitive information beyond what isolated images expose.
  Although life-logging video streams can substantially improve utility of these promising systems, they also introduce significant privacy risks by revealing sensitive information, such as behavioral patterns, emotional states, and social interactions, beyond what isolated images expose.
  If unresolved, these risks may undermine public trust and hinder the sustainable development of always-on AI technologies. Existing privacy protections are either attack-specific or incur substantial utility loss,
  % , leaving no whole-process solution for open-world deployment.
  and fail to consider the entire data exploitation pipeline.
  We therefore posit that the privacy-utility trade-off in life-logging video streams is a foundational challenge for next-generation AI systems that demands further investigation. We call for novel pipeline-aware privacy-preserving designs that jointly optimize utility and privacy for long-horizon life-logging visual data. In parallel, formal privacy leakage metrics and standardized benchmarks remain important open directions for future research.
\end{abstract}

\section{Introduction}
% \tianyuan{}
% \sijie{}
% \yl{}

Recent rapid advances in proactive agents~\cite{openclaw_official_website,deng2025proactive,deng2024towards,zhang2024ask}, surging attention in Vision-Language Models (VLMs)~\cite{simeoni2025dinov3,carion2025sam,bolya2025perception} and world model development~\cite{assran2023self,deepmind_genie3_world_models_2025}, collectively suggest a broader transition of AI systems from episodic, prompt-driven tools toward persistent agents that continuously perceive, remember, and act in the physical world.
Rather than passively responding to explicit user queries, emerging systems are increasingly designed to actively interact with users' real world situations through cameras~\cite{meta_rayban_smart_glasses},  microphones~\cite{rayneo_official_website}, wearable sensors~\cite{yang2025proagent}, and long-term memory modules~\cite{openclaw_official_website}.
This shift can already be seen in various products and research directions, from always-on assistants in smart glasses by Meta~\cite{meta_rayban_smart_glasses}, Google~\cite{google_android_xr_gemini_glasses_2025}, RayNeo~\cite{rayneo_official_website}, HeyCyan~\cite{heycyan_official_website}, and always-on AI agent prototypes like VisionClaw~\cite{liu2026visionclaw}, 
% which have positioned always-available multimodal assistants as a key interface for everyday assistance, 
% advocate always-on egocentric perception for opportunistic task execution
% to proactive anomaly surveillance systems built on ~\cite{jeon2024pass}
to daily life assistant with proactive home-assistant robots~\cite{yamasaki2025multi},
% \tianyuan{search for more examples?} 
All of them rely on persistent sensing of the environment, especially the visual environment, to provide persistent assistance.
These trends clearly indicate that continuous observation, especially the capture and exploitation of long-horizon (life-logging) vision information, is becoming a core design principle of next-generation intelligent systems.

Significant utility improvement can be achieved with this persistent sensing trend, as shown in previous work that diffusion models trained on video data consistently outperform their image-only counterparts on various downstream tasks~\cite{velez2025image}. Proactive agents~\cite{zhou2026mem1}, personalized assistants~\cite{hong2026tameing}, adaptive robotics~\cite{wang2026memex} all benefit from long-horizon reasoning that leverages rich temporal information in interaction histories and continuous environment perception, rather than relying on isolated prompts.
% This shift is supported by rapid \textit{\textbf{always-on}} hardware progress and real-world deployment: wearable devices like lightweight smart glasses~\cite{meta_rayban_smart_glasses,google_android_xr_gemini_glasses_2025,rayneo_official_website} and body-worn cameras~\cite{dji_nano_official_page,looki_l1_official_product_page,axon_body_cameras_complete_guide} now provide affordable, energy-efficient platforms for persistent sensing in everyday environments; meanwhile, the widespread of \textit{\textbf{always-on}} home security systems~\cite{vivint_official_website,simplisafe_value_home_security}, public surveillance cameras~\cite{spot_ai_security_guard,earthcam_explosionprotected_streamcam,earthcam_explosionprotected_streamcam_robotic} and car dash cameras~\cite{iiwey_4channel_dashcam_collection,neideso_official_website} have already made life-logging longitudinal visual streams accessible, providing the physical substrate for continuous personal assistance and proactive intelligence.
This shift is not only supported by the widespread deployment of \textit{\textbf{always-on}} visual recording systems, including home security systems~\cite{vivint_official_website,simplisafe_value_home_security}, public surveillance cameras~\cite{spot_ai_security_guard,earthcam_explosionprotected_streamcam,earthcam_explosionprotected_streamcam_robotic}, and car dash cameras~\cite{iiwey_4channel_dashcam_collection,neideso_official_website}, which have already made longitudinal life-logging visual streams accessible; but also further advanced by the rapid advances in \textit{\textbf{always-on}} hardware, including wearable devices such as lightweight smart glasses~\cite{meta_rayban_smart_glasses,google_android_xr_gemini_glasses_2025,rayneo_official_website} and body-worn cameras~\cite{dji_nano_official_page,looki_l1_official_product_page,axon_body_cameras_complete_guide}, which provides affordable, energy-efficient platforms for persistent sensing in everyday environments, and enriches the physical substrate for continuous personal assistance and proactive intelligence.

% Yet the same always-on cameras and other sensors that boost utility simultaneousness bring unprecedented privacy exposure:
% Yet, since limited on-device (cameras or sensors) resources~\cite{shao2025ai} often require the use of remote servers for improved training and inference therefore prevent data from remaining strictly on-device, the same always-on devices that boost utility also increase privacy exposure during sharing and processing.
Yet, as limited on-device (user-side camera, sensor and digital device) resources~\cite{shao2025ai} often require the use of remote servers for improved training and inference, preventing data from remaining strictly on-device, the same always-on systems and devices that boost utility also increase privacy exposure during sharing and processing.
Beyond revealing identity~\cite{wang2021deep,dhar2021pass} or biological information~\cite{jain2022intelligent} that an isolated image can disclose,
longitudinal video streams expose richer signals like personal geographical information~\cite{zhang2025pervasive}, behavior patterns~\cite{li2019collaborative,ulutan2020actor}, emotional states~\cite{legara_2023_frame_emotion_video}, etc., that no single frame could disclose. 
Crucially, privacy risk accumulates over time: seemingly harmless information (e.g. habitual routines) can become highly sensitive when aggregated over time.
Moreover, always-on sensing threatens not only primary users but also bystanders who may be recorded without awareness or consent.
% If these threats remain unaddressed, they may erode public trust, and even significantly hinder the long-term adoption and sustainable development of AI technologies, not limiting to always-on AIs. Legal regulations such as GDPR~\cite{gdpr2016}, CCPA~\cite{ccpa2018}, and HIPAA~\cite{hipaa1996} (in the healthcare domain) have also underscore these concerns from a legal perspective.
If unaddressed, these privacy risks could erode public trust and hinder the broader adoption of AI, as reflected by regulations such as GDPR~\cite{gdpr2016}, CCPA~\cite{ccpa2018}, and HIPAA~\cite{hipaa1996}.

However, existing privacy-preserving methods failed to meet pipeline-aware privacy protection in open-ended environments without prior knowledge of what kind of privacy threats the system will face. Deleting raw data or blocking them from exploitation is the most direct privacy preservation mechanism, but harms utility to the most. Other protection methods like face de-identification~\cite{gafni2019live}, pose anonymization~\cite{huang2025recoverable} and audio-video cycle-VQ-VAE~\cite{xu2021audio} are either attack-specific or setting-specific and failed to generalize to other potential privacy threats and settings, while other more general protection methods like pixel-level blurring~\cite{bhutani2025preserving,wang2025privacy} or perturbation\cite{machanavajjhala2017protecting}, anonymization~\cite{brkic2017know,more2024privacy}, stego visual data creation~\cite{chen2018cartoongan,meng2019steganography}, Differential Privacy (DP)~\cite{dwork2006differential}, Federated Learning (FL)~\cite{mcmahan2016federated} 
% trades utility for better privacy to a large extent.
achieve stronger privacy by largely sacrificing utility and are not pipeline-aware.

Therefore, we argue that,
as AI systems inevitably move toward exploiting life-logging visual streams, confronting the privacy-utility trade-off becomes a foundational challenge rather than an option. Without adequate protection, easy disclosure of personal information undermines public trust.
Thus, for the sustainable development of AI system, developing a pipeline-aware privacy-preserving system, from data collection to exploitation, with good privacy-utility trade-off for life-logging video stream is of urgent need. 
% These life-logging video streams necessitate confronting privacy-utility trade-offs as a vital research and product invention problem. 
% Addressing this challenge requires system-level designs considering privacy-utility trade-off balance. Promising directions include not only privacy preservation methods, but also formal privacy leakage metrics and benchmarks that jointly measure utility and privacy risk.
This calls for system-level solutions, including not only privacy-preserving high-utility methods, but also formal metrics and benchmarks to jointly evaluate utility and privacy risk.

% In the following, we first summarize the existing privacy risks in life-logging video stream systems to demonstrate the intensive privacy threat inherit therein in \cref{sec:privacy_threat}, and then summarize and analyze why existing defense mechanisms are not adequate to protect against such privacy threats in \cref{sec:defense}. We move on to explicitly list the requirements of a good defense mechanism in \cref{sec:good_defense} and give a possible development path to show that the proposed defense requirements are implementable in \cref{sec:possible_path} with basic evaluation principles open for improvements.
In the following, we first summarize the privacy risks in life-logging video stream systems to highlight their inherent threats (\cref{sec:privacy_threat}), and then analyze why existing privacy protections are insufficient (\cref{sec:defense}). We then outline the requirements of a good privacy-preserving mechanism (\cref{sec:good_defense}) and present a preliminary feasible development path with preliminary evaluation principles (\cref{sec:possible_path}).

\section{Escalated Privacy Issue with Life-Logging Video Streams}\label{sec:privacy_threat}

% \begin{figure}
%     \centering
%     \includegraphics[width=0.5\linewidth]{}
%     \caption{Caption}
%     \label{fig:placeholder}
% \end{figure}
\begin{figure}[!tb]
    \centering
    \subfigure[]{\includegraphics[width=0.24\textwidth]{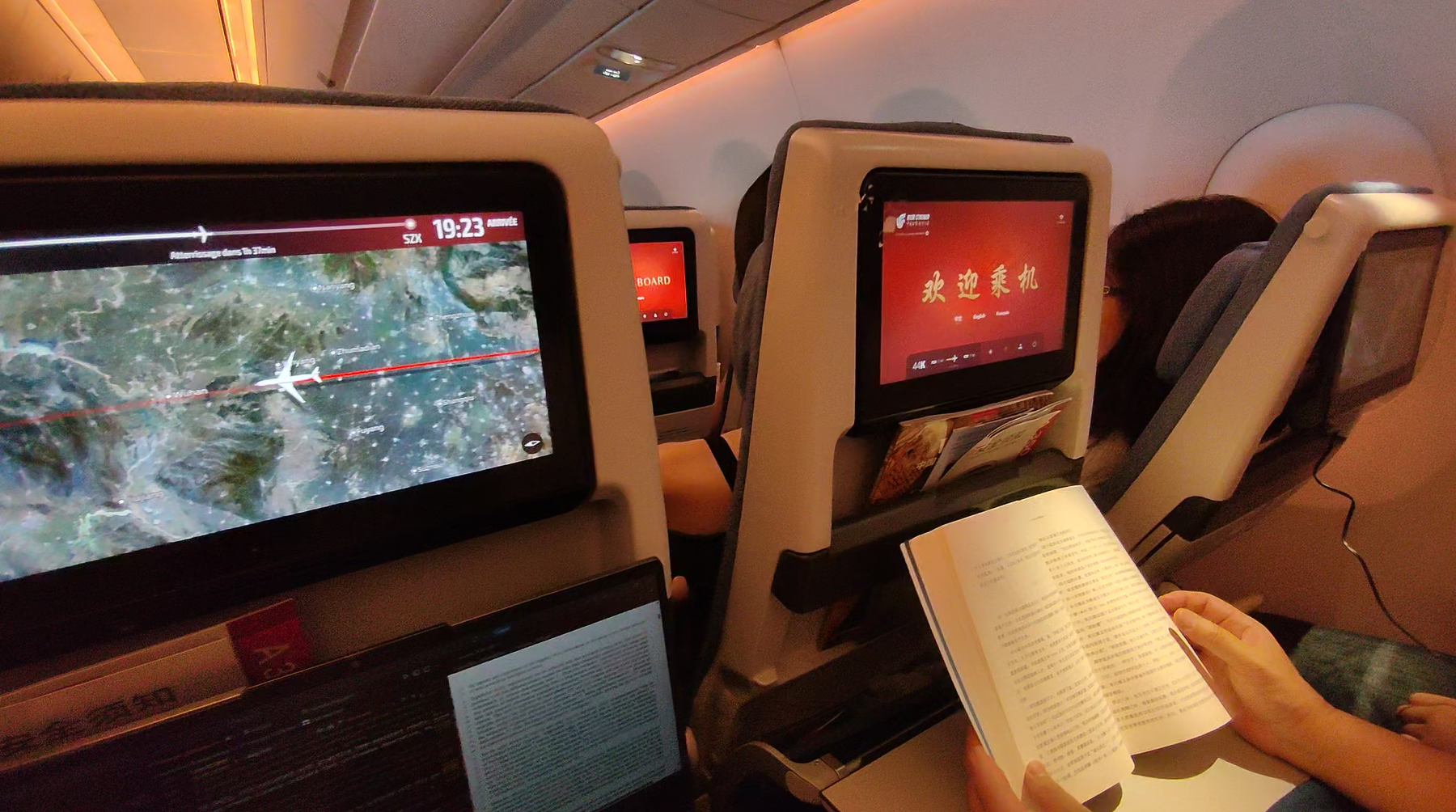}} 
    \subfigure[]{\includegraphics[width=0.24\textwidth]{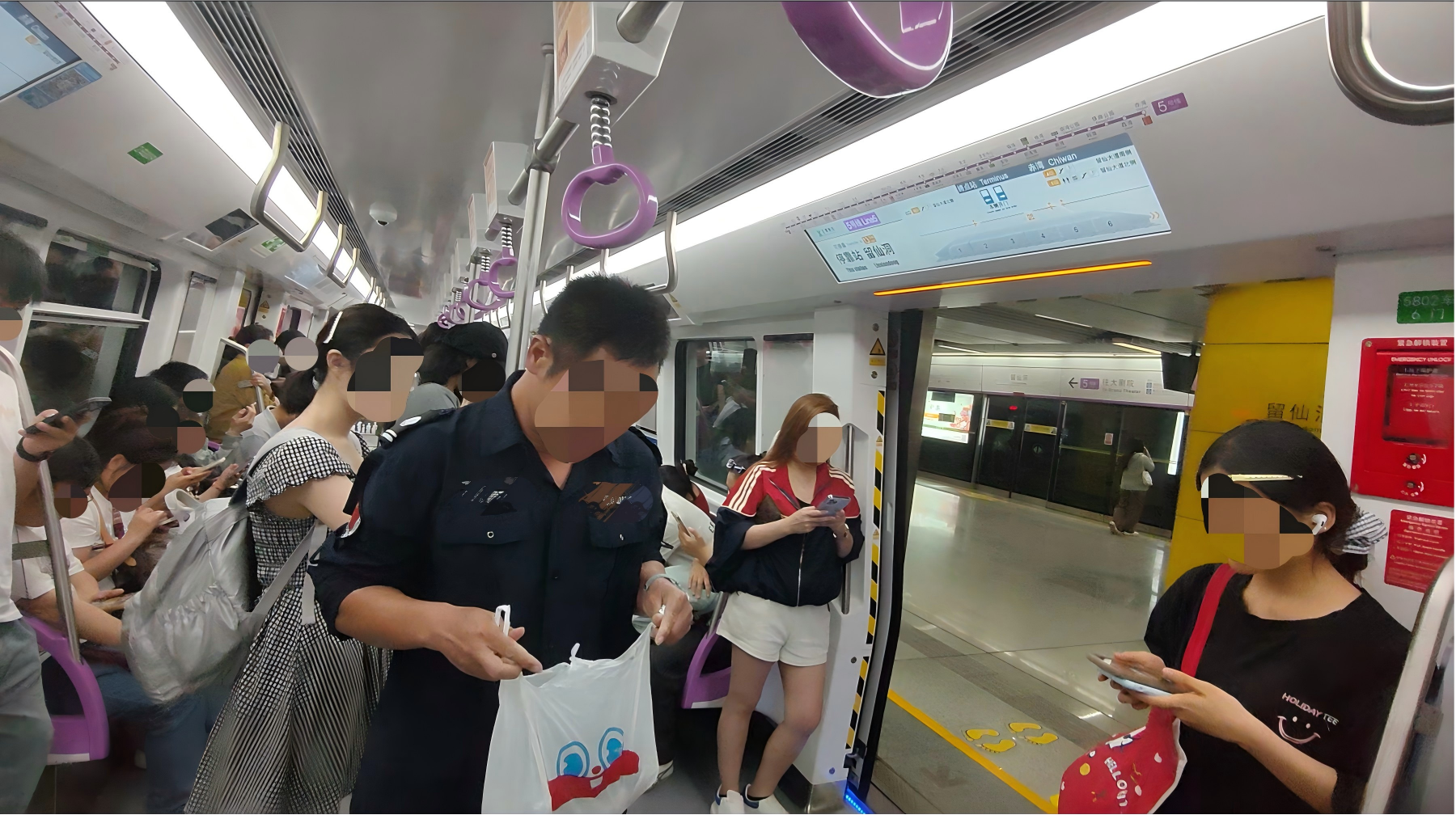}} 
    \subfigure[]{\includegraphics[width=0.24\textwidth]{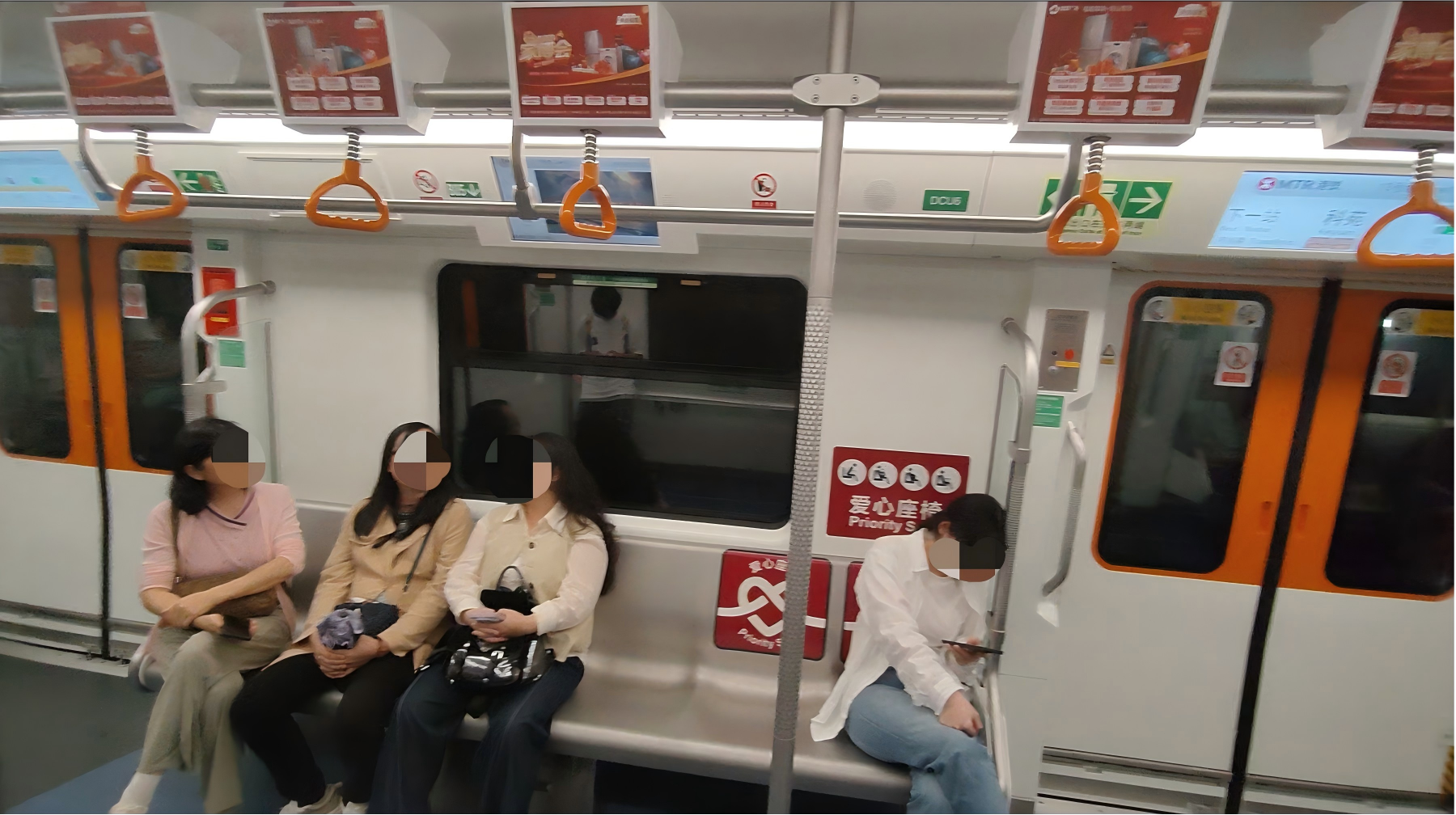}}
    \subfigure[]{\includegraphics[width=0.24\textwidth]{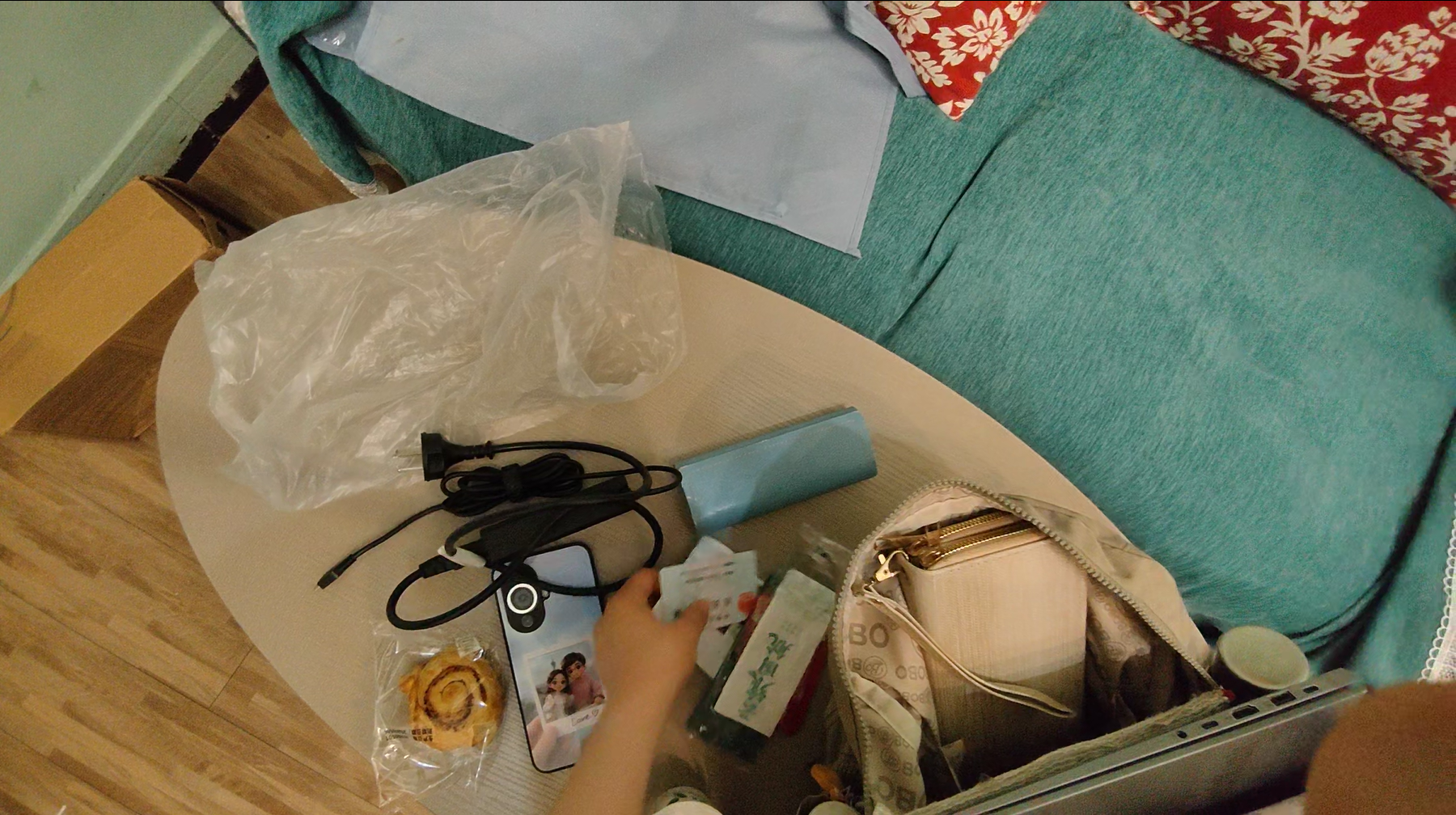}}
    % \subfigure[]{\includegraphics[width=0.32\textwidth]{figures/smart_glasses/location.png}} 
    % \subfigure[]{\includegraphics[width=0.32\textwidth]{figures/smart_glasses/pedestrian-2.png}} 
    % \subfigure[]{\includegraphics[width=0.32\textwidth]{figures/smart_glasses/relation-gender.png}}
\caption{Frames with privacy risks illustrated: (a) user’s whereabouts; (b) bystanders’ facial identities and locations; (c) bystanders’ identities and their relationships; (d) user’s relationship status and potential gender. Captured by RayNeo-V3 smart-glasses. For ethical considerations, bystanders’ facial features and occupational information are blurred .}
% \caption{Frames containing obvious privacy risks. Captured by RayNeo-V3 smart-glasses. \tianyuan{@sijie, is this okey?} Privacy risks illustrated: (a) reveals the user’s whereabouts and identity (the seat number of the user can be inferred); (b) reveals bystanders’ identities and their relationships; (c) infers the user’s relationship status and potential gender.}
\label{fig:rayneo_privacy_recording}
\vspace{-1em}
\end{figure}

Data insufficiency for model training (especially for world understanding) and personalized agent development point to the collection of continuous user daily data (life-logging data). 
The rise of always-on hardware, like portable cameras, smart glasses, and surveillance systems, establishes an ideal physical foundation for large-scale collection of life-logging data.
However, while bringing a promising image of improving the performance of current vision models, these data record and document our most precious moments which at the same time are very likely to be highly private. \cref{fig:rayneo_privacy_recording} shows some examples that contain various types of privacy risks. % If misused intentively, they can reveal various private information posing greate privacy threat.

In the following, we first summarize when and how privacy threats arise throughout the workflow of collecting and exploiting life-logging data with \textit{\textbf{always-on}} devices in \cref{subsec:privacy_risks_on_workflow}. We then examine these risks in detail for visual data in \cref{subsec:raw_data_privacy,subsec:non_raw_data_privacy}, with particular emphasis on how video streams amplify them. Privacy threats based on raw data and non-raw data are summarized in \cref{subsec:raw_data_privacy} and \cref{subsec:non_raw_data_privacy} respectively.

\subsection{Privacy Risks To be Protected} \label{subsec:privacy_risks_on_workflow}

\begin{figure}[!tb]
\centering
    \includegraphics[width=0.99\linewidth]{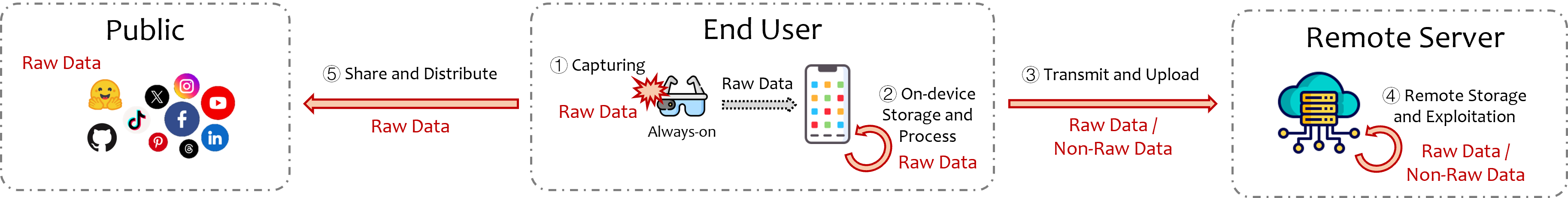}
\caption{Demonstration of how data flaw across the life-logging video stream workflow.}
\label{fig:data_pipeline}
\vspace{-1em}
\end{figure}
% \todo{Put this before 2.1 \& 2.2}
% \todo{introduce raw data and embedding here}

% \tianyuan{Use brain storm things to enrich}

In this part, we summarize and categorize potential privacy threats in the entire life-logging video stream collection and utilization workflow (see \cref{fig:data_pipeline}). 
% We assume a setting where the recordings are captured by edge devices with limited computational resources which does not support on-device model training. Therefore, given the precious knowledge contained in these data, they have to be uploaded to remote servers for further analysis and exploitation (e.g. vision model training). This is a very practical setting that is broadly applied in real-world applications~\cite{shao2025ai}. 
Here, we consider a practical setting where resource-limited edge devices cannot support on-device training or inference of large powerful models~\cite{shao2025ai}, requiring data upload to remote servers for better analysis and exploitation~\cite{shao2025ai}.
In the following, we categorize and summarize privacy threats according to whether they are based on raw data or not, with detailed privacy threats summarized in \cref{subsec:raw_data_privacy,subsec:non_raw_data_privacy} respectively.

% % \textit{\textbf{Recording.}} Raw data is captured by cameras and/or sensors. Since no analysis is performed in this process, we can assume that no privacy risk arises. \tianyuan{???}
% \textit{\textbf{Recording, On-device Storage and Process.}} At first glance, after raw data capturing by end devices (cameras and/or sensors), user-side on-device storage of data is safe. However, this may not be the case for individuals who are unconscious of their existence in the recording, such as pedestrians on the road~\cite{zhang2025more}. Adversarial end users can infer sensitive personal information, like those summarized in \cref{subsec:raw_data_privacy}, of these appeared individuals using their intentionally recorded raw vision data. 
% % However, these are not risks introduced by persistent recording but appear with the invention of recording devices. 
% % We believe that as long as recording devices give end users the choice when to start and when to stop as well as what to delete, the responsibility for eliminating these risks should be attributed to each end user as they are the one who actively records these data. 
\textit{\textbf{Recording.}} 
% At first glance, after raw data capturing by end devices (cameras and/or sensors), user-side on-device storage of data is safe. However, this may not be the case for individuals who are unconscious of their existence in the recording, such as pedestrians on the road~\cite{zhang2025more}. Adversarial end users can infer sensitive personal information, like those summarized in \cref{subsec:raw_data_privacy}, of these appeared individuals using their intentionally recorded raw vision data. 
In this process, end users capture their desired raw data by cameras and/or sensors. End users (humans) is able to extract sensitive information mentioned in \cref{subsec:raw_data_privacy} from these raw data.
% With these raw data, all privacy risks summarized in \cref{subsec:raw_data_privacy} exit, even only ``human process'' is done.
% However, these are not risks introduced by persistent recording but appear with the invention of recording devices. 
% We believe that as long as recording devices give end users the choice when to start and when to stop as well as what to delete, the responsibility for eliminating these risks should be attributed to each end user as they are the one who actively records these data. 

\textit{\textbf{On-device Storage and Process.}} In this part, captured raw data remain on the user-side. At first glance,
% after raw data capturing by end devices, 
% user-side on-device storage and process of data 
this process
is safe. However, this does not hold for individuals unaware of being recorded, such as pedestrians on the road~\cite{zhang2025more}. 
% Adversarial end users can infer sensitive personal information, like those summarized in \cref{subsec:raw_data_privacy}, of these appeared individuals using their intentionally recorded raw vision data. 
If maliciously exploited, sensitive personal information, such as those summarized in \cref{subsec:raw_data_privacy}, of these appeared individuals can be inferred from the raw data.

\textit{\textbf{Transmit and Upload.}} We focus on uploading data to remote servers in this part. 
% If raw data are transmitted and shared, as mentioned in \cref{subsec:raw_data_privacy}, various kinds of recorded individuals' sensitive private information can be directly leaked, including identity, behavior pattern, location information, etc. 
Transmitting raw data, as mentioned in \cref{subsec:raw_data_privacy}, directly exposes private information of end users and recorded individuals such as identity, behavior pattern and location information.
Moreover, even encrypted metadata sharing~\cite{li2016side} or data embedding sharing remain unsafe, as illustrated in \cref{subsec:non_raw_data_privacy}. 
% Therefore, we argue that a \textbf{privacy-preserving representation} of the raw data should be established and served as the form of information that is used for transmission.

\textit{\textbf{Remote Storage and Exploitation.}} 
% After transmission and upload, data kept on remote server also contain potential privacy risks, not only during storage but also during exploitation for model training. Either raw data or vision models trained using these raw data are stored, all attacks mentioned in \cref{subsec:raw_data_privacy,subsec:non_raw_data_privacy} can all be launched respectively if intended, which can even be further enhanced with cross-modal and temporal information available.
After transmission and upload, data on remote servers still pose significant privacy risks during storage, training, and inference. Whether raw data or vision models trained on them are retained, the attacks outlined in \cref{subsec:raw_data_privacy} or \cref{subsec:non_raw_data_privacy} remain applicable, and may be further amplified by the availability of cross-modal and accumulated temporal information.

\textit{\textbf{Share and Distribute.}}
This part considers sharing data to the public, such as sharing on social media or open-source as public datasets. As these data are very likely to be raw data, all privacy risks mentioned in \cref{subsec:raw_data_privacy} exist. 
% However, we argue that, as sharing and distribution are both user-initiated behaviors, publishers should take over the privacy protection responsibility.

\subsection{Intense Privacy Threats from Exposing Raw Visual Data} \label{subsec:raw_data_privacy}
Exposing raw visual data is highly risky.
\textit{Identity recognition} is the most direct visual threat, extracting sensitive content such as facial features~\cite{wang2021deep} and facial attributes~\cite{dhar2021pass} from seemingly mundane pictures, 
with recognition accuracy exceeding $99\%$~\cite{he2016deep} using Convolutional Neural Networks (CNNs). 
% Studies have shown that the accuracy of face recognition systems based on Convolutional Neural Networks (CNNs) on standard datasets has exceeded $99\%$~\cite{he2016deep}. 
Pretrained Vision-Language Models (VLMs), like CLIP, can also be exploited to realize cross-platform identity recognition and tracking
% by extracting facial feature vectors from 
using images~\cite{radford2021learning}. Moreover, \textit{re-identification (Re-ID)}, an extension to \textit{identity recognition} threat, can further transform the identification knowledge to new images~\cite{li2023clip,zheng2016person}. Video-based \textit{identity recognition}~\cite{aggarwal2004system} and \textit{Re-ID}~\cite{zheng2016person,pathak2020video} have long been explored,
% by the academic community, 
with richer vision information further aiding temporal linkage~\cite{pathak2020video} compared to images.

Not only can identify information be inferred from raw data, but also sensitive static \textit{biometric information} of users, like fingerprints~\cite{maltoni2009handbook,jain2022intelligent}, irises~\cite{kaur2014review}, to name a few~\cite{millett2010biometric}, can be revealed merely from images as well. 
When temporal information is further available with videos, 
% in addition to static biological information, 
\textit{dynamic physiological features} like gait~\cite{zhu2021gait}, gaze~\cite{tran2020you} and region of interest (ROI) on face~\cite{bhutani2025preserving} can be further analyzed.
These kinds of unique and somewhat persistent information, once secretly analyzed and used by adversaries, are likely to pose a permanent threat to privacy (even safety), such as biological information abuse, gaining advantages in negotiations and analyzing health status.

More private information can be leaked with raw video data. For example, based on spatiotemporal models, \textit{actions and behavior of individual or group} can also be recognized from videos~\cite{li2019collaborative,ulutan2020actor,vellenga2026ai,wang2026hybrid,asali2026spatiotemporal}, enabling in-depth deduce of information such as individuals' and groups' activity intentions, social roles and relationships, and professional characteristics~\cite{wang2026hybrid}. 
Action detection can be further performed when combined with other modality, like audio~\cite{shi2021multi}, a natural combination with video.
This information can be maliciously used to monitor or supervise specific individuals or groups.

Moreover, facial (micro-)expressions and body languages can be exploited to infer individuals' \textit{emotional states, psychological characteristics, and personality traits} from raw video data. Large pretrained VLMs are superior and mature in such tasks~\cite{legara_2023_frame_emotion_video} with products already introduced to the market and users~\cite{morphcast_emotion_video,imentiv_video_emotion_recognition}. Combining with audio modality further aids emotional analysis~\cite{wang2025multimodal} from video.
If the marketing or human resource management team intentionally accesses relevant information, they can be more guiding or aggressive, affecting or misleading clients' real need.

Last but not least, VLMs are even better than humans in analyzing \textit{geographical location} from videos, inferring where the video is recorded~\cite{zhang2025pervasive}, since they are better at aggregating subtle information from different temporal frames. 
This information can be maliciously used for military reconnaissance or espionage, posing great threat on personal safety.

\subsection{Privacy Threats Persist with Processed (Non-Raw) Visual Data} \label{subsec:non_raw_data_privacy}

% Previously, a simple yet intuitive idea is that as long as private data are kept locally at end-user side, their privacy is guaranteed as raw data is never exposed to other parties. 
However, what makes the situation intense is that simply restricting raw data to on-device storage is necessary but fundamentally insufficient. 
% Various modern inversion attacks are designed to operate on intermediate representations of trained vision encoders~\cite{hintersdorf2024does} and vision models~\cite{li2024membership,hu2025membership}, side-information~\cite{li2016side} or even vision model update gradients~\cite{hatamizadeh2022gradvit}. These attacks are able to reconstruct training data sample, posing great threat on training data privacy. 
Various attacks are designed to operate on intermediate representations of trained vision encoders~\cite{hintersdorf2024does}, vision models~\cite{li2024membership,hu2025membership}, side-information~\cite{li2016side}, update gradients~\cite{hatamizadeh2022gradvit,fang2023gifd}, or even adversarial samples~\cite{xue2026privacy}. 
% To make it more severe, privacy risks amplify with temporal linkage and multi-modal fusion. 
We elaborate on them in detail in the following.

\begin{figure}
\centering
  \subfigure[]{\includegraphics[width=0.48\textwidth]{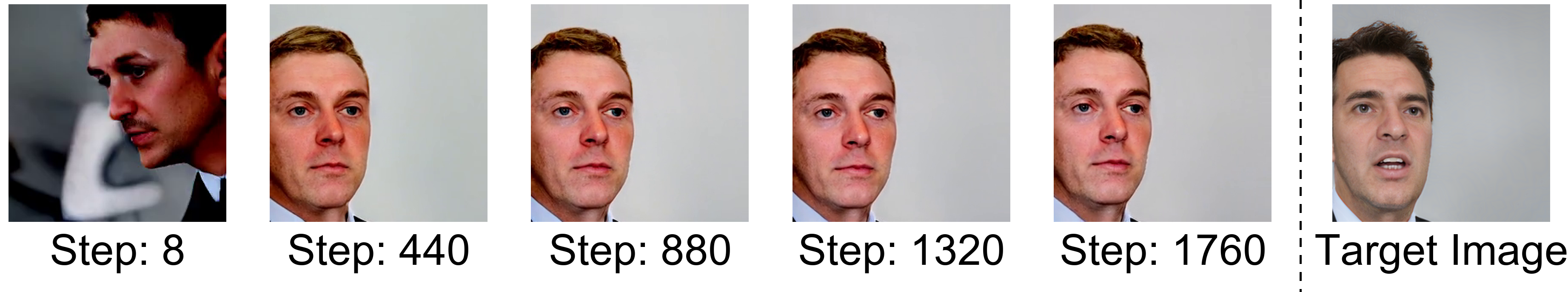}}
  \hspace{0.4em}
  \subfigure[]{\includegraphics[width=0.48\textwidth]{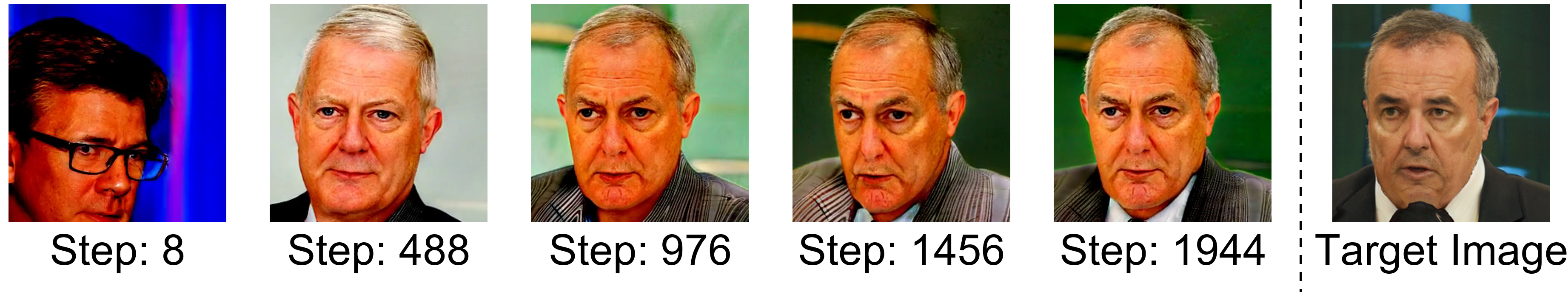}}
\caption{Human face inversion attack based on embeddings from PE~\cite{bolya2025perception}. Facial features are restored.}
\label{fig:main_inversion_attack}
\vspace{-1em}
\end{figure}

\textit{Membership Inference Attack (MIA)} is kind of often-seen privacy threat brought by vision models, 
% Instead of analyzing and extracting sensitive information from inference time presented images, this privacy threat aims to steal by reconstructing training time image data (or senstivie information). 
aiming to steal private information by reconstructing (training time) original image data or sensitive information. 
Various kinds of vision models, including powerful pretrained VLMs such as CLIP~\cite{hintersdorf2024does} and LlaVA~\cite{li2024membership,hu2025membership}, as well as cutting edge vision encoders like SigLIP~\cite{zhai2023sigmoid,tschannen2025siglip}, PE~\cite{bolya2025perception} and SAM~\cite{carion2025sam}, can be exploited by attackers to extract original images from embeddings with only access to black-box model (only model API and output embeddings are available)~\cite{allakhverdov2026feature}, also testified by our local results shown in \cref{fig:main_inversion_attack} with details included in \cref{sec:appendix_inversion_attack}. With richer information from video~\cite{li2025vidsme,wang2026vidleaks}, MIA achieves an AUC higher than $90\%$ on particular tasks~\cite{wang2026vidleaks}, escalating the privacy threat. When combined with other modality like text (which can be extracted from audio), \textit{sensitive features of training data}, like label and caption information, can also be successfully inferred with high recovery rate~\cite{xiu2025caprecover}.

\textit{Gradient Inversion Attack (GIA)} inverts raw training data as well but relying on model update gradients instead of model outputs~\cite{hatamizadeh2022gradvit,fang2023gifd}. GIA assumes a white-box setting, where the victim model is accessible by the attacker. Vision data can further aid other modality, like text, in GIA, achieving fuzzy string matching ratio~\cite{sharmageetika_fuzzy_string_matching} higher than $90\%$~\cite{hemo2025gradient}. 

\textit{Side-Channel Attack (SCA)} aims to infer recorded individual's sensitive attributes like activity or behavior from third party eavesdropping. For example, Activity of Daily Living (ADL) can even be extracted from encrypted video stream for transmission~\cite{li2016side}, using only transmission spikes.

Note that since our focus is data privacy, we omit the discussion on adversarial attacks that aims to modify model behaviors in this part.

% \section{Limitations of Existing Privacy-Preserving Methods}\label{sec:defense}
\section{Existing Privacy Protections Fall Short for Life-Logging Video Streams}\label{sec:defense}

In this section, we first (\cref{subsec:privacy_protection_responsibility_requirement}) clarify our arguments on privacy-preservation requirements and responsibility allocation within each process mentioned in \cref{subsec:privacy_risks_on_workflow} and \cref{fig:data_pipeline}, and then summarize representative relevant privacy-preserving mechanisms (\cref{subsec:limitations_of_defenses}) to show that none of them already meets the need in the entire life-logging video stream workflow.

\subsection{Privacy Preservation Requirements and Responsibility Allocation} \label{subsec:privacy_protection_responsibility_requirement}
In the following, we analyze the privacy preservation requirements and allocate privacy preservation responsibility to end users, recording device providers and platforms, remote severs, etc.

% \textit{\textbf{Recording, On-device Storage and Process.}} \todo{separate to 2 part}
% We posit that when recording devices provide end users with explicit control over when to initiate, terminate, and delete recordings, the primary responsibility for privacy protection should reside with the end users themselves. This is because they are the active agents who determine when and what data is captured. 
% % However, this assumption becomes increasingly fragile in always-on and automated sensing scenarios, where user awareness and control may be limited or absent.
\textit{\textbf{Recording.}}
We posit that, as recording devices provide end users with explicit control over when to start, stop, and delete recordings, the primary responsibility for privacy protection should reside with end users. This is because they are the active agents who determine what data is captured.\footnote{Following legal regulations like GDPR, explicit consent from recorded bystanders should be achieved. Thus, for recording device providers, user alters can be a choice if improper recording (highly sensitive or even against law) is detected on-device.}
% However, this assumption becomes increasingly fragile in always-on and automated sensing scenarios, where user awareness and control may be limited or absent.

\textit{\textbf{On-device Storage and Process.}} 
% We posit that, even on-device storage and process need to be done, recording device and platform providers should strictly adhere to protocols like never sharing raw data outside the edge device and only perform required storage, access and process by end users.
We posit that, when on-device storage and processing are necessary, recording device providers and platforms should strictly adhere to basic privacy-preserving protocols including: \emph{\underline{raw data should never}} \emph{\underline{leave the edge device}}, and any storage, access, or processing should be limited to what is explicitly required and authorized by the end user.\footnote{From a safety alignment perspective, on-device processing unit should refuse end users' malicious capturing or inference and misuse of the data. However, such safety concerns fall outside the privacy focus of this paper, therefore not our focus.}
% However, this assumption becomes increasingly fragile in always-on and automated sensing scenarios, where user awareness and control may be limited or absent.

\textit{\textbf{Transmit and Upload.}}
We posit that an \emph{\underline{attack-agnostic privacy-preserving representation}} of raw data should be established as the sole form of information permitted for transmission. 
% Moreover, as there is no prior knowledge on what kind of privacy risks will the system needs to face, the representation should not be designed for specific attacks, but rather a more general privacy-preserving one. 
Attack-agnostic is essential under unbounded open-world privacy threats (even under fixed application cases), requiring such representations to defend against diverse attacks.
Recording device providers and platforms are responsible for designing and deploying mechanisms to generate such representations on end-devices, as they control the data transmission pipeline, protocol, and the underlying infrastructure.

\textit{\textbf{Remote Storage and Exploitation.}} 
We posit that the included storage, training and inference processes should also satisfy privacy requirements in this part. At the same time, effective exploitation demands high utility. Therefore, recording device providers and platforms, or the remote server, should be responsible for designing and deploying mechanisms that produce \emph{\underline{attack-agnostic privacy-preserving high-utility representations}} of the raw data, which guarantees the privacy of not only the exploit representations but also the model trained using these representations.

\textit{\textbf{Share and Distribute.}}
We posit that 
% , similar to the first case, 
sharing and distribution are user-initiated actions. Therefore, end users, as publishers, should assume the primary responsibility for privacy protection.

\subsection{Limitations of Existing Privacy-Preserving Methods} \label{subsec:limitations_of_defenses}
From \cref{subsec:privacy_protection_responsibility_requirement}, we argue that the privacy-preserving goal is to find an attack-agnostic protection mechanism that has a good trade-off between privacy and utility and can be directly utilized for training without the need of decoding or reconstructing the raw data as a pre-process. 

The ``attack-agnostic'' requirement rules out privacy-preserving methods designed for specific threats, such as face de-identification~\cite{gafni2019live} and QM-VAE~\cite{qiu2022novel} for face recognition threat, reversible anonymization~\cite{huang2025recoverable} for pose estimation threat, cycle-VQ-VAE~\cite{xu2021audio} for audio-visual settings, etc. We therefore focus on more general privacy-preserving methods that can defend against diverse attacks.

% \tianyuan{Should we introduce the semi-honest and adversarial term in this paper?}
% Given all kinds of semi-honest (the attacker only wants to infer sensitive information from exposed data) and adversarial (the attacker aims to maliciously change the model behavior) attacks in vision modality, various kinds of defense methods have been proposed in literature, inspired by previous work~\cite{liu2024vertical,zou2024vflair} which can be generally separated into general defense methods that helps to prevent multiple kinds of privacy threats, as well as specific defense methods targeted for particular kind of privacy threats or attacks.

% \subsection{General Defense Methods}

\textit{Disguising real visual information}, either by \textit{pixel-level blurring}~\cite{wang2025privacy,bhutani2025preserving} or by \textit{pixel-level perturbation} through adding noise~\cite{machanavajjhala2017protecting}, is a very direct way of preventing adversaries from inferring sensitive private information (e.g. face, hairstyle, cloth color, biometric attributions,
% can be hidden under the added noise
etc.~\cite{brkic2017face}) from raw visual data. However, these methods trade utility to large extent for higher privacy~\cite{machanavajjhala2017protecting}.

\textit{Anonymization} is another approach to protect data privacy. Generative adversarial networks (GANs) are often trained to synthesize images that \textit{replace or modify sensitive regions}, such as identity-bearing faces~\cite{brkic2017know,hukkelaas2019deepprivacy,hellmann2024ganonymization,more2024privacy}, human bodies~\cite{brkic2017know}, and moving objects~\cite{uittenbogaard2019privacy}.
% , or \textit{performs style transformation}~\cite{chen2018cartoongan} to protect certain level privacy. 
Denoising AutoEncoder~\cite{machanavajjhala2017protecting} has also been explored for similar purposes. 
Although these methods achieve high visual realism and preserve downstream utility, they are difficult to train~\cite{fu2026contrastive}, requiring careful loss design, and often struggle with temporal consistency in video, such as cross-frame face alignment~\cite{more2024privacy}.

\textit{Steganography} disguises the whole raw visual data to protect data privacy through GAN-based steganography to create a stego image~\cite{tang2017automatic,meng2019steganography}, style transformation~\cite{chen2018cartoongan,wu2019secgan}, or optimal noise injection~\cite{kim2019latent}. However, these methods are either hard to train~\cite{wu2021two}, or having the possibility of reconstructing the original image~\cite{fu2020secure}, or failed to provide a good privacy-utility trade-off~\cite{zhang2019invisible}.

% \tianyuan{Add other GAN utilized setting given by \url{https://chatgpt.com/c/69d4ce20-73c4-83e8-907e-6957186042b7}.}
% \tianyuan{Read this ``Investigating Vulnerabilities and Defenses Against Audio-Visual Attacks: A Comprehensive Survey Emphasizing Multimodal Models'' and ``Privacy-Preserving Video Anomaly Detection: A Survey''}

\textit{Encryption}, such as Homomorphic Encryption (HE)~\cite{gentry2009fully} and Secure Multi-Party Computation (MPC)~\cite{yang2019federatedML} and function encryption~\cite{xu2021fedv}, 
% % successfully protects data transmission safety and data analysis by third-party with encrypted data, and is widely applied in real world application~\cite{li2016side,??}\tianyuan{citation}. Although 
% although effective in preventing direct access of third-party to raw data during transmission, various kinds of side-channel information leakage~\cite{li2016side,li2020your,huang2023rethinking,yetim2026tracking} have proven to be highly effective. Sensitive information including users' 
% % activities of the daily living (ADL) 
% activity pattern~\cite{li2016side,li2020your,huang2023rethinking}, behavior habit~\cite{huang2023rethinking}, and geospatial information~\cite{yetim2026tracking} captured by the video can still be inferred from encrypted video streams through eavesdropping on the wireless transmissions, exploiting merely encrypted video stream and specific transmission traffic pattern.
prevents direct access to raw data during transmission. However, side-channel leakage attacks~\cite{li2016side,li2020your,huang2023rethinking,yetim2026tracking} have shown that sensitive information, including user activity patterns~\cite{li2016side,li2020your,huang2023rethinking}, behavioral habits~\cite{huang2023rethinking}, and geospatial information~\cite{yetim2026tracking}, can still be inferred from encrypted video streams by eavesdropping on wireless transmissions, without accessing the underlying plain content.

\textit{Differential Privacy (DP)}~\cite{dwork2006differential,wang2020videodp} provides a mathematic privacy-preserving guarantee and disguises each individual samples in the whole population. However, it inherently suffers from bad privacy-utility trade-off~\cite{zou2024vflair}, since its underlying mechanism is random noise injection~\cite{dwork2006differential}.
% \subsection{Emerging Specialized Defense Methods}

\textit{Federated Learning (FL)}~\cite{mcmahan2016federated,yang2019federatedML} is a kind of distributed learning paradigm that aims to protect training data privacy without the need of centralizing each participant's local data. Instead, only model updates (gradients) or intermediate results are shared~\cite{yang2019federatedML}. However, FL still suffers from various of data leakage risks~\cite{ratnayake2023review,zou2024vflair}, and is often combined with DP to seek further privacy preservation~\cite{shi2021hfl,noble2022differentially} therefore suffering from achieving high privacy while maintaining good utility.

% Reversible anonymization~\cite{huang2025recoverable} has been studied for Human Pose Estimation (HPE) by jointly optimizing a privacy enhancing module, a privacy recovery module, and a pose estimator for maintaining HPE utility while preserving privacy. With in-sensor privacy enhancement, malicious attacker cannot infer sensitive personal information from disguised image while authorized party like police still has the ability to recover real image. 

% ~\cite{bhutani2025preserving} tested defending methods including multiple blurring operations, various noise addition strategies, and different time-averaging techniques to defend against remote
% photoplethysmography (rPPG) which effectively estimates users’ physiological parameters.

% Reversible anonymization~\cite{huang2025recoverable} has been studied for Human Pose Estimation (HPE)
% ~\cite{gafni2019live} proposed face de-identification methods for video-based face recognition

Note that, we omit the discussion on privacy-preserving methods by \textit{access control}~\cite{schaffer2005video} since they directly harms the utility of the data from the training perspective.

\subsection{
Lack of Pipeline-Aware Privacy Protection Throughout the Entire Workflow
}
Aside from their inability to meet the requirements analyzed in \cref{subsec:privacy_protection_responsibility_requirement}, existing methods also do not provide a pipeline-aware solution for processing raw visual data throughout the entire data pipeline, from capture and transmission, to storage, inference and model training.

% Furthermore, current defense methods are typically tailored to narrow visual scenarios~\cite{}, with little empirical evidence supporting their generalization to more diverse or realistic settings. 
% These limitations become even more pronounced in diverse real-world settings, where data exhibits substantial variability across environments, users, and privacy threat types.
% \tianyuan{Add example from user study and defense paper.}

Therefore, we argue that in the emerging era of always-on (visual) recording, a \textit{\textbf{pipeline-aware, attack-agnostic, privacy-preserving, high-utility mechanism}} is necessary to mitigate inherent privacy risks. Such a systematic design is essential for the sustainable development of the field of always-on AI techniques and systems, as it enables users to trust and permit the use of their data in consistently advancing more capable vision models and systems.

% privacy-preserving systems must be inherently adaptive and capable of handling heterogeneous scenarios and addressing privacy threats that vary both in type and severity.

\section{What is A Good Privacy Solution for Life-logging Video Stream Systems}\label{sec:good_defense}

\subsection{Essential Design Objectives} \label{subsec:design_goal}

% To define what attributes a good always-on visual privacy protection technique or system should possess, it is crucial to first clarify the ultimate objective. Based on our review of prior work and analysis of real-world recorded data, we identify the following essential qualities that a robust always-on privacy solution should exhibit:
To guide the design of effective always-on visual privacy solutions, it is essential to first define the core design objectives. Drawing on prior research and real-world recordings we gathered ourselves, we outline the major goals that a well-performing privacy protection system should achieve.

\textit{\textbf{Goal 1:} Preserving general downstream utility is as important as protecting user privacy.} Always-on visual data is inherently valuable for improving user services, models and products on various downstream tasks. Privacy mechanisms must therefore avoid excessive degradation of any possible downstream task performance while enforcing strong privacy guarantees.

\textit{\textbf{Goal 2:} Privacy protection should generalize across diverse attacks in various real-world settings.} 
Privacy requirements vary drastically across scenarios. For instance, data from private bedrooms often requires protection of sensitive content, such as nudity, whereas in public dining spaces, the focus is on safeguarding individuals’ identities. Effective solutions should be to adapt to such heterogeneous contexts and differing threat models.

\textit{\textbf{Goal 3:} End users should always possess control over their proper original data.} 
While privacy protections should be enforced at the product provider and platform or remote server side, end users must retain access and control to their unaltered proper (e.g. not violating laws) data. This is essential for both user agency and practical needs such as legal requirement or investigative use.

\subsection{Attributes of A Good Privacy-Preserving System} \label{subsec:good_attributes}

Based on the above analysis, we further list the attributes that we believe a good privacy preservation system should possess as follows:

\textbf{Attribute 1:} Real data should be kept only at end users devices and should never be exposed to remove server or company.

\textbf{Attribute 2:} After on-device privatization, only privacy-preserved information is transmitted to the server side for storage and possible further exploitation.

\textbf{Attribute 3:} Transmitted privacy-preserved information should preserve various downstream model training and task performance while preventing reconstruction of the original raw data, other possible privacy leakages as well as adversarial attacks.

% \textbf{Attribute 4:} Transmitted privacy-preserved information and the whole system are both robust against adversarial attacks.

% Once a system satisfies all the above attributes, for involving parties other than the end user, with the having no information on how raw data privatization is done, the transmitted, stored, and further exploit information is original data non-reconstructable. Therefore, no privacy is exposed. 

\subsection{Evaluation Principles} \label{subsec:evaluation_principle}

The evaluation of such systems should be structured along the following key dimensions. Within each dimension, practitioners can flexibly instantiate specific downstream tasks, sensitive attributes, attack models, and system constraints, allowing application-specific evaluation while maintaining a unified framework. 

\textbf{Downstream Task Performance (DTP).} 
% This metric evaluates system utility by jointly assessing performance across a diverse set of downstream tasks of interest, providing a comprehensive measure of overall effectiveness.
This metric measures system utility by jointly evaluating performance across a set of downstream tasks of interest, providing a comprehensive assessment of effectiveness and utility.

\textbf{Privacy Leakage (PL).} 
% This encompasses evaluating a wide range of sensitive attribute inference attack success rate, not only at the single-video or frame (image) level, but also through cross-video and cross-image linkage, as well as cross-modal linkage, where information is aggregated to amplify privacy risks.
This quantifies the success of inferring sensitive attribute of interest, encompassing not only single-video or single-image (i.e. single-frame) scenarios, but also cross-video, cross-image, and cross-modal linkage, where aggregated information can amplify privacy risks.

\textbf{Reconstruction Resistance (RR).} 
% This assess to what extent possible adversaries can reconstruct the original data. All sorts of known attacks can be added and removed regarding applicant's interest.
This evaluates the extent to which adversaries can reconstruct the original input from the representation which is undesirable. This dimension supports a flexible suite of reconstruction attacks, allowing practitioners to tailor evaluation to relevant threat models.

\textbf{System Efficiency (SE).} 
% This evaluation covers all sorts of computational and communication cost, for example, privatization latency, privatization computational cost, privacy-preserved information transmission cost, etc. Applicants can freely select what should be considered and evaluated referring to their consideration.
This captures computational and communication overhead, including privatization latency, on-device computation cost, transmission cost of privacy-preserved information, etc. Specific metrics can be adapted to reflect deployment constraints.

\section{An Implementable Preliminary Design of the Pipeline-Aware Privacy-Preserving System}\label{sec:possible_path}

In this section, we give one possible implementable preliminary design of the \textit{\textbf{pipeline-aware privacy-preserving system}} for always-on systems. This system encompass a privatization representation construction module for obtaining the \textit{Minimal Sufficient Representation (\textbf{MSR})} of the give raw data, which could be pretrained for once and then distributed for various use, and a communication module following restricted protocols controlling the flow of sensitive information.

% Non-Reconstructable Representation (NRR) that satisfies the following is good ...

% Useful but Unreconstructable

\subsection{On-device Privatized Representation Construction Module}

Visual models evolve into general-purpose systems supporting \emph{various downstream tasks}. % , they inherently enable a wide range of intended and unintended inferences. 
% In such open-ended settings, privacy risks can no longer be captured by a fixed set of predefined sensitive attributes (e.g., identity or gender), as \emph{future tasks} and \emph{emergent inference capabilities} are not known a prior. This renders task-specific protection fundamentally insufficient. Meanwhile, general pixel-level perturbation or blurring harms struggles to achieve a good privacy-utility trade-off. 
In such open-ended settings, downstream tasks cannot be exhaustively enumerated as a fixed set of targets, nor can privacy risks be characterized by a predefined list of sensitive attributes, rendering task-specific protections inherently insufficient. Meanwhile, general pixel-level obfuscation (e.g., blurring or perturbation) often fails to achieve a satisfactory privacy–utility trade-off.
% In such open-ended settings, not only downstream tasks cannot be listed as a fixed list of targets, but also privacy risks cannot be depicted by a fixed set of predefined sensitive attributes, rendering task-specific protections insufficient. On the other hand, general pixel-level obfuscation (e.g. blurring or perturbation) often fails to achieve a satisfactory privacy–utility trade-off.

We therefore argue that a promising solution for balancing the privacy–utility trade-off is to seek the \textit{minimal sufficient} representation, denoted by $z$, of each raw input $x$ with respect to all possible tasks $Y$ in the given application setting via multi-objective optimization. In pure statistical term, $z$ is the \textit{minimal sufficient statistics} of $x$ with respect to $Y$, which is closely aligned with the \textit{Information Bottleneck Principle}~\cite{tishby2000information,tishby2015deep}. Also, instead of trying to find a pixel-level privatized result, privacy can be enforced at the \textit{\textbf{representation level}}, i.e. the dimension of $z$ does not need to be limited to the same as that of $x$. Therefore, we name $z$ as the \emph{\textbf{Minimal Sufficient Representation (MSR)}} of $x$. The above can be formulated as:
\begin{equation} \label{eq:information_bottle_neck_guarantee}
\max I(z; Y) \quad \text{s.t.} \quad I(z; x) \leq \epsilon,
\end{equation}
where $I(\cdot,\cdot)$ denotes the \textit{Mutual Information}~\cite{shannon1948mathematical} between the random variables and $\epsilon$ controls privacy protection strength. By optimizing \cref{eq:information_bottle_neck_guarantee}, we force $z$ to capture the least relevant features of $x$ while removing irrelevant parts that do not contribute to the task $Y$.  

On the privacy side, by directly constraining $I(z; x)$, we treat all residual information as potentially sensitive, eliminating the need to enumerate private attributes. We simultaneously eliminate the possibility for high fidelity reconstruction, since none-relevant information is dismissed and therefore cannot be reconstructed. On the utility side, enumerating all possible downstream tasks $Y$ is impractical, as we stated previously. One possible solution is to maximize $I(z, x_{semantic})$ instead of maximizing $I(z; Y)$, where $x_{semantic}$ is the ``semantic content'' of $x$ that provides all the required information for possible $Y$. In practice, $x_{semantic}$ can be achieved by using powerful pretrained foundational encoders, like PE~\cite{bolya2025perception}, DINOv3~\cite{simeoni2025dinov3}, SAM 3~\cite{carion2025sam} and etc., for the vision modality. Or, one can train from scratch using contrastive learning, masked modeling, etc.

To sum up, a privatized representation construction model $\mathcal{M}$, that outputs privatized MSR $z$ of a given $x$, can be trained with the following objective:
\begin{equation} \label{eq:M}
\min\limits_{z}  \left\{- I(z; x_{sentiment}) + \lambda I(z; x) \right\} = \min\limits_{\Phi_{\mathcal{M}}} \left\{- I(\mathcal{M}(x); x_{sentiment}) + \lambda I(\mathcal{M}(x); x) \right\},
\end{equation}
where $\Phi_{\mathcal{M}}$ is the trainable parameters of $\mathcal{M}$. In practice, $I(z; x_{sentiment})$ can be replaced by all sorts of utility loss function, like distance measurement function $\mathcal{L}_{d}(z, x_{semantic})$ when $x_{semantic}$ is produced by pretrained foundational encoders, or InfoNCE loss~\cite{oord2018representation} if trained from scratch. Off-the-shelf techniques such as Variational Information Bottleneck (VIB)~\cite{alemi2017deep}, which leverages the reparameterization trick, can be directly applied to transform $I(z; x)$ into a tractable and differentiable objective, and have been shown to be effective~\cite{zou2024defending}.

The goal of such a procedure is to train the representation model $\mathcal{M}$ to suppress signals exploitable by adversaries, effectively providing protection against privacy ricks. Meanwhile, general-purpose utility is explicitly preserved by integrating contrastive or self-supervised objectives ($I(z,x_{sentiment})$), ensuring that $z$ captures adequate amount of semantics for various downstream tasks. Note that this is a principled approach that does not need pre-specification of targeted downstream tasks or sensitive privacy attributes.

We need to clarify that this privatization module should be deployed on edge devices at user-side to satisfy \textbf{Attribute 1} given in \cref{subsec:good_attributes}. \textbf{Attribute 3}
% and \textbf{Attribute 4} 
given therein is also satisfied.

\subsection{Privatized Representation Communication Module}
Since edge devices at user-side, like camera sensors, smart glasses, autonomous vehicles, mobile phones, to list a few, are often of limited resources~\cite{shao2025ai}, remote servers are often introduced to support more powerful models for better task performance. Meanwhile, remote servers can also serve as the aggregation node to help gain a more comprehensive view through either information aggregation or federated training to unify these divergent data to obtain a more powerful model. 

Under such setting, to satisfy \textbf{Attribute 2} given in \cref{subsec:good_attributes}, we argue that only the privatized representation as well as the user query or prompt should be transmitted to the remote server, whereas merely answer and response from remote server model should be transmitted back to user side device.

% \subsection{Wide Application Setting}
% \tianyuan{???}

\subsection{Discussions on Concerns about the Design} \label{subsec:concerns_of_proposed_method}

% One may argue that as long as performance of downstream tasks, such as face recognition, is maintained, user privacy is inevitably leaked. 
% % However, we clarify that privacy leakage is not equivalent to task performance, but rather to the amount of extraneous information about the raw input that remains accessible from the representation. 
% However, we clarify that privacy leakage is not determined by whether a downstream task succeeds, but by what additional information beyond the task requirement can be extracted from the privatized representation. For example, a face recognition system only requires features that distinguish identities (e.g., relative facial structure), but a rich embedding may also encode texture, skin condition or other side information, which is irrelevant to recognition yet sensitive. 
A common concern is that preserving performance on downstream tasks, such as face recognition, necessarily implies privacy leakage. But we clarify that, privacy leakage should be assessed not by task success, but by the amount of \emph{task-irrelevant information} retained in the process. 
For example, face recognition fundamentally relies on identity-discriminative features (e.g., relative facial structure), yet learned embeddings often capture additional details such as texture, skin condition, or other side information that are irrelevant to the task but privacy-sensitive.
However, under the lens of our proposed MSR, an ideal representation preserves only the information necessary for downstream task generalization while discarding all redundant content, eliminating the concern.

Another concern is the absence of formal theoretical privacy guarantees for the proposed MSR, such as explicit mathematical bounds quantifying privacy protection like what is done in \textit{Differential Privacy (DP)}~\cite{dwork2006differential}. We view this not as a flaw, but as an open research challenge. In particular, establishing a principled link between \textit{Mutual Information} and \textit{DP} may offer a viable pathway toward mathematical rigorous privacy guarantees.

\subsection{Difficulty and Risk for Implementation} \label{subsec:difficulty_of_proposed_method}

\textbf{Imperfect surrogate objectives for $I(z;Y)$ may degrade downstream generalization.} To avoid enumerating all downstream tasks, we approximate $I(z;Y)$ using proxies such as $I(z; x_{sentiment})$, where $x_{sentiment}$ is derived from pretrained encoders or learned from scratch. However, the former can be incomplete and biased, while the latter can be difficult to optimize. Consequently, they may omit some task-relevant information, potentially undermining downstream utility and generalization.\footnote{
% Another suboptimal surrogate solution might be listing several most vital downstream tasks for a fixed given application setting and directly optimize their task performs in the objective function. This method is promising in guarantee the selected downstream task performance but may fall short on other unselected tasks.
Another possible (suboptimal) surrogate approach is to specify several key downstream tasks as $Y$ for a given application and directly optimize their performance in the objective function. While this may preserve performance on the selected tasks, it may generalize poorly to unselected ones.
}

\textbf{Multi-objective training for privacy-utility balancing is inherently difficult.}
% The multi-objective training goal formularized in \cref{eq:information_bottle_neck_guarantee} is inherently hard to obtain. 
% Following previous work on Information Bottleneck~\cite{alemi2017deep}, we regard the privatized representation construction process as a Markov chain $Y-x-z$. Therefore, following the Data Processing Inequality (DPI)~\cite{beaudry2012intuitive}, we have $I(z; Y) \leq I(z; x)$. Therefore, \cref{eq:information_bottle_neck_guarantee} remains a privacy-utility trade-off objective.
Following prior Information Bottleneck work~\cite{alemi2017deep}, we regard the privatized representation construction process as a Markov chain $Y-x-z$. By the Data Processing Inequality (DPI)~\cite{beaudry2012intuitive}, $I(z;Y)\leq I(z;x)$, making the objective in \cref{eq:information_bottle_neck_guarantee} fundamentally a trade-off between conflicting privacy and utility goals.
While high utility requires preserving information, privacy demands removing it.
Therefore, the multi-objective training goal formularized in \cref{eq:information_bottle_neck_guarantee} is inherently hard to obtain. 
Careful design of what objective functions to be used and how to combine them (e.g. scaler for loss balance), as well as how to stabilized training are all unexplored questions.
% \textbf{Overly disguise could lead to utility collapse.}
% Following previous work on Information Bottleneck~\cite{alemi2017deep}, we regard the privatized representation construction process as a Markov chain $Y-x-z$.
% Therefore, following the Data Processing Inequality (DPI)~\cite{beaudry2012intuitive}, we have
% % $I(Y; x) \geq I(z; Y)$ and 
% $I(z; Y) \leq I(z; x)$. Therefore, \cref{eq:information_bottle_neck_guarantee} remains a privacy-utility trade-off objective. Therefore, overly emphasizing on the privacy part will eventually lead to utility collapse on downstream task, Therefore, investigating useful techniques that helps better balance this trade-off is vital for real-world deployment.

\textbf{Temporal accumulation could lead to new privacy threat.}
One deep concern is that, with privatized representation accumulated overtime, will temporal information leak more privacy compared to having merely respective data sources? This is still an unexplored question in the always-on setting.

\textbf{Model evolution could lead to embedding failed to generalizable.}
Another concern is whether the evolution of vision models for downstream tasks will result in pre-generated $z$ becoming less useful. How to construct an encoder model $\mathcal{M}$ that does not evade overtime and whether it is possible to construct an adapter when there exists the need to convert the current privatized representation to align with new encoding are both interesting research questions.

\section{Conclusions and Call to Actions}\label{sec:conclusion}

In this paper, \textbf{we highlight the severe privacy risks emerging in the always-on era with life-logging video streams}, which is further amplified by temporal linkage and multi-modal aggregation. 
% If left unaddressed and still chasing behind merely utility improvement, these privacy threats may lead to long-term societal harm: 
If left unaddressed in the pursuit of utility, these privacy threats may have long-term harm to individuals, undermine sustainable AI development, and even ultimately threaten societal stability:
% The pervasive collection of visual data, capturing behaviors across both public and private spaces, can make individuals feel constantly observed, recorded, and profiled by sensor providers and large-scale AI systems. When such data is further used for model training, it raises (at least) the unsettling prospect of AI systems implicitly “remembering” individuals. This erosion of privacy can ultimately undermine public trust in AI. 
life-logging visual data collection can make individuals feel constantly observed and unsafe; when used for training, it raises concerns about AI systems ``remembering`` individuals, eroding public trust.
% We argue that overlooking these challenges is detrimental not only to individuals, but also to the sustainable development and deployment of AI systems, and may ultimately threaten broader societal stability. 
Therefore, % privacy risks inherent in always-on recording must be treated as a central concern.
\textbf{we advocate that these privacy risks require urgent attention on confronting the privacy-utility trade-off issue under the always-on setting}. We posit that, aside from utility pursuit, robust privacy protection is fundamental to human-centered AI and essential for the long-term sustainability and societal acceptance of AI systems.

% \textbf{We further present a preliminary implementable pipeline-aware framework to mitigate the aforementioned privacy risks}. However, our design remains largely conceptual. Practical training and deployment strategies require further investigation. We warmly welcome alternative and advanced solutions for systematic comparison and collaborative advancement. In addition, exploring previously unrecognized privacy risks, as well as developing rigorous benchmarks to evaluate privacy–utility trade-offs, remain urgent and open research directions.
% Further, 
\textbf{We take a first step by outlining a preliminary pipeline-aware framework for mitigating these risks.} Yet this is only a starting point. Practical training, deployment, and validation remain open, and no single approach will suffice. We call for competing and polished solutions for principled comparisons, and collective progress. Equally urgent is the need to surface previously unrecognized privacy risks and to establish rigorous benchmarks for evaluating privacy–utility trade-offs.

Addressing these challenges requires action from the research ecosystem to support always-on device providers. Venues such as NeurIPS and other leading AI conferences can play a key role by prioritizing privacy-preserving techniques, standardized benchmarks for life-logging video streams, and by fostering dialogues between technical and policy communities. These efforts are essential to ensure reliable, sustainable, and socially aligned AI development.

% \section*{Acknowledgments}
% An Acknowledgments section, if used, \textbf{immediately precedes} the References. Sponsorship information and funding data are included here. The preferred spelling of the word ``acknowledgment'' in American English is without the ``e'' after the ``g.'' Avoid expressions such as ``One of us (S.B.A.) would like to thank\ldots'' Instead, write ``F.~A.~Author thanks\ldots'' Sponsor and financial support acknowledgments are also to be listed in the ``acknowledgments'' section.
% \tianyuan{??}

\bibliography{ref}

% An Appendix, if needed, should appear before the acknowledgments.
\newpage
\appendix
\section*{Appendix}

\section{Inversion Attack on Non-Private Image Embeddings Produced by Pretrained Encoders} \label{sec:appendix_inversion_attack}
% \yl{I don't know the exact name of ``Inversion Attack''. I just see the word above and use it. Just change it if it's improper.}
% Figure~\ref{fig:inversion_attack}

We consider a black-box setting where the goal is to reconstruct the original image from embeddings produced by a cutting-edge pretrained image encoder. We show experimentally that such non-private embeddings, learned without explicit privacy constraints, can leak sensitive information about the original private information (the input), revealing inherent privacy risks. This observation reinforces our claim that,even under a strict on-device raw data constraint, storing non-private pretrained models or their embeddings (representations) on remote servers remains fundamentally privacy-sensitive.

% Note that since our goal is not designing a new inversion attack for better image reconstruction performance, but instead to prove that training data can be leaked even with cutting edge black-box pretrained image encoders, which shows that even storing the model trained without privacy protection will leak private information. Therefore, we keep the attack design and implementation simple as long as it is effective for demonstration.
Note that our goal is not to design a novel inversion attack to improve reconstruction quality, but rather to demonstrate that original input data can be leaked from cutting-edge pretrained image encoders in a black-box setting for highlighting that VMs trained without privacy considerations can themselves become sources of privacy leakage. Consequently, we adopt a simple attack design and implementation, provided that it is sufficient to demonstrate this privacy threat.

\subsection{Inversion Attack Methodology}

We formulate the inversion attack as reconstructing the original image $\mathbf{I}_{gt}$ from its embedding $\mathbf{e}_{gt}$ produced by the attack-targeted encoder$\mathcal{E}$:
$$\mathbf{e}_{gt} = \mathcal{E}(\mathbf{I}_{gt}).$$
To this end, we leverage a pretrained image generator $\mathcal{G}$ that maps latent codes $\mathbf{w}$ to images:
$$\mathbf{I} = \mathcal{G}(\mathbf{w}).$$
The attack is then formulated as an optimization problem over $\mathbf{w}$, seeking a generated image whose embedding ($\mathcal{E}(\mathcal{G}(\mathbf{w}))$) matches the target embedding ($\mathbf{e}_{gt}$):
$$ \mathbf{e} = \mathcal{E}(\mathcal{G}(\mathbf{w})) \approx \mathbf{e}_{gt}, $$
or equivalently,
$$ \mathbf{w}^* = \arg\min_{\mathbf{w}} \mathcal{L}_{\mathcal{E}}(\mathcal{E}(\mathcal{G}(\mathbf{w})), \mathbf{e}_{gt}),$$
where $\mathcal{L}_{\mathcal{E}}$ measures the discrepancy between the embeddings. We solve this optimization via gradient descent, iteratively updating
$$ \mathbf{w} \leftarrow \mathbf{w} - \eta \nabla_w \mathcal{L}_{\mathcal{E}} = \mathbf{w} - \eta \frac{\partial \mathcal{L}_{\mathcal{E}}}{\partial \mathbf{w}}, $$
where $\eta$ is the learning rate. This procedure yields an optimized latent code $\mathbf{w}^*$ whose generated image approximates the original input in the embedding space.

\subsection{Implementation Detail} \label{subsec:appendix_inversion_attack_implementation}

The encoder $\mathcal{E}$ under attack is the Perception Encoder (PE)~\cite{bolya2025perception}, a cutting-edge powerful pretrained vision encoder that extracts rich semantic representations from images, producing embeddings of size $1025 \times 192$ per image. For the generator $\mathcal{G}$, we use StyleGAN3~\cite{karras2021alias} trained on their locally assembled FFHQ-U and the AFHQv2 dataset for the inversion of human faces and animals respectively\footnote{Model weights ``stylegan3-r-ffhqu-256x256.pkl'' for human faces and ``stylegan3-r-afhqv2-512x512.pkl'' for animals are downloaded from  \url{https://catalog.ngc.nvidia.com/orgs/nvidia/teams/research/models/stylegan3/files?version=1}.}, with image resolution $256 \times 256$ and $512 \times 512$ for human faces and animals respectively and latent codes ($\mathbf{w}$) dimension $512$ for both.\footnote{The choice of StyleGAN3 is motivated by its compatibility with recent PyTorch versions (version 2.8.0), aligning with the requirements of the attacked encoder $\mathcal{E}$ (PE). However, a key limitation is its bias toward generating animal-centric images due to its animal-only training data. As a result, it is less effective at inverting images from other domains, such as human daily activities, which constrains the scope of our privacy leakage from pretrained image encoder embeddings demonstration. Nevertheless, we anticipate that more powerful generators trained on larger and more diverse datasets would enable similarly effective inversion across embeddings without privacy protection (like that given by PE) of a wider range of image types.}

We adopt mean squared error (MSE) as the loss function $\mathcal{L}_\mathcal{E}$ and optimize the latent code $\mathbf{w}$ via gradient-based optimization. The learning rate is initialized at $1.0$ and linearly decayed to $0.95$ over $2000$ iterations. Each experiment is carried out on a single $80$G-A$100$ GPU.

To align with the generative capacity of $\mathcal{G}$, the target images $\mathbf{I}_{gt}$ are synthesized by $\mathcal{G}$ from randomly sampled latent codes $\mathbf{w}_{gt}$. During reconstruction, neither $\mathbf{I}_{gt}$ nor $\mathbf{w}_{gt}$ is accessible to the attacker; only the target embedding $\mathbf{e}_{gt} = \mathcal{E}(\mathbf{I}_{gt})$ is available on the attacker side.
% , we generate $16$ images with the same StyleGAN3 generator as the target to be attacked: we set 16 random seeds (from $0$ to $15$) and generate 16 noises as the ground truth latent code $\mathbf{w}_{gt}$ and then generate the corresponding image ($\mathbf{I}_{gt}$). 

During the reconstruction process, for each target, we initialize $16$ random latent codes ($16$ random noises) using random seeds from $100$ to $400$ (step size $20$) and perform reconstruction from each initialization. For evaluation, we record not only the optimization loss (embedding discrepancy $\mathcal{L}$), but also the pixel-level difference $\mathcal{L}_\mathbf{I} := ||\mathbf{I} - \mathbf{I}_{gt}||_2$ and the latent difference $\mathcal{L}_\mathbf{w} := ||\mathbf{w} - \mathbf{w}_{gt}||_2$, which are not used during optimization. The final reconstruction is selected based on the minimal embedding discrepancy, regardless of whether it is located in the middle of or in the end the optimization process.

\begin{figure}[!tbp]
  \centering

  % \begin{minipage}{0.99\linewidth}
  %   \centering
  %   \includegraphics[width=\linewidth]{figures/liangyue/draft/280_19_pe/sum.png}
  % \end{minipage}

  % \begin{minipage}{0.99\linewidth}
  %   \centering
  %   \includegraphics[width=\linewidth]{figures/liangyue/draft/380_29_pe/sum.png}
  %   \\[-0.25em]
  % \end{minipage}

  \begin{minipage}{0.99\linewidth}
    \centering
    \includegraphics[width=\linewidth]{figures/liangyue/draft/280_18_pe/sum_larger.png}
  \end{minipage}

  \begin{minipage}{0.99\linewidth}
    \centering
    \includegraphics[width=\linewidth]{figures/liangyue/draft/300_22_pe/sum_larger.png}
    \\[-0.25em]
  \end{minipage}
  
  \begin{minipage}{0.99\linewidth}
    \centering
    \includegraphics[width=\linewidth]{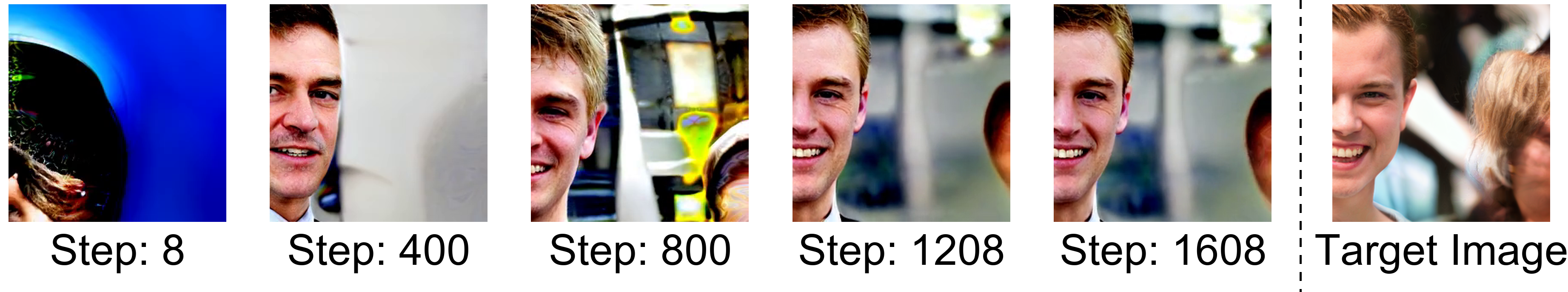}
    \\[-0.25em]
  \end{minipage}

  \begin{minipage}{0.99\linewidth}
    \centering
    \includegraphics[width=\linewidth]{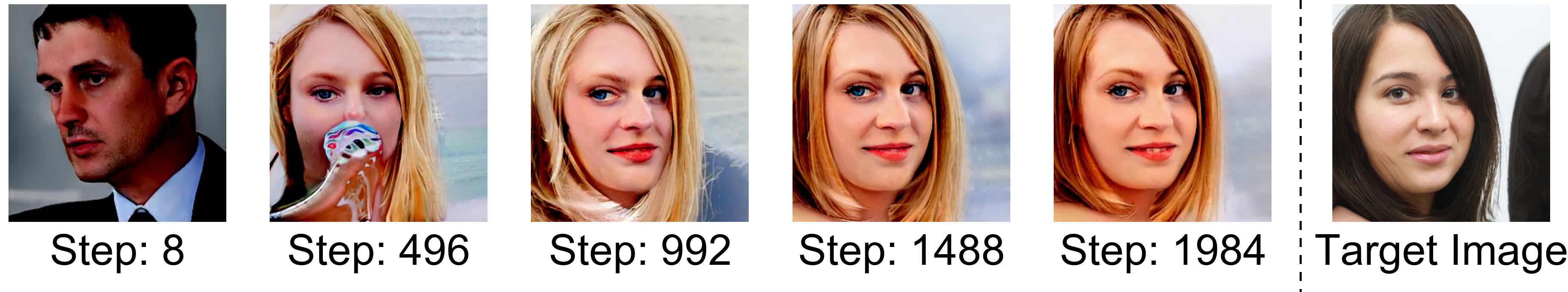}
    \\[-0.25em]
  \end{minipage}

  \caption{(Human faces) Inversion attack reconstruction results of the original image from embeddings produced by a cutting-edge black-box pretrained image encoder PE~\cite{bolya2025perception}.}
  \label{fig:inversion_attack_human}
\end{figure}

\begin{figure}[!tbp]
  \centering

  \begin{minipage}{0.99\linewidth}
    \centering
    \includegraphics[width=\linewidth]{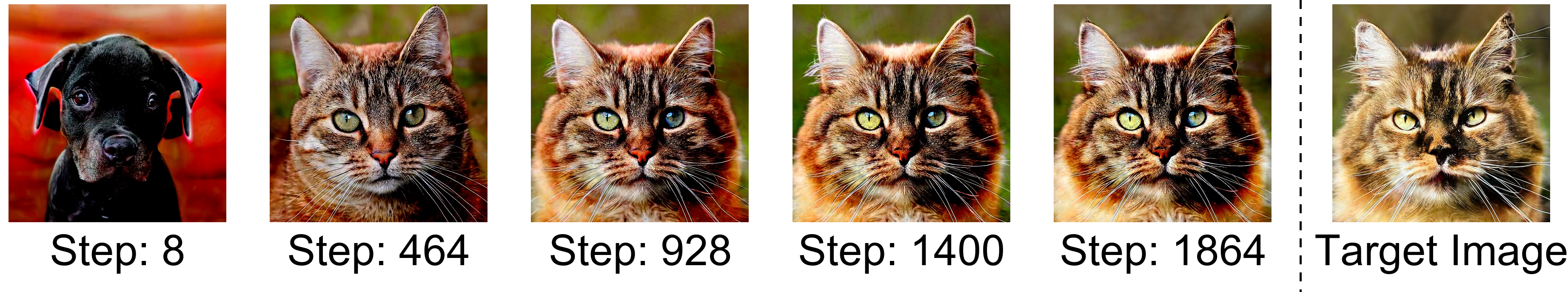}
  \end{minipage}

  \begin{minipage}{0.99\linewidth}
    \centering
    \includegraphics[width=\linewidth]{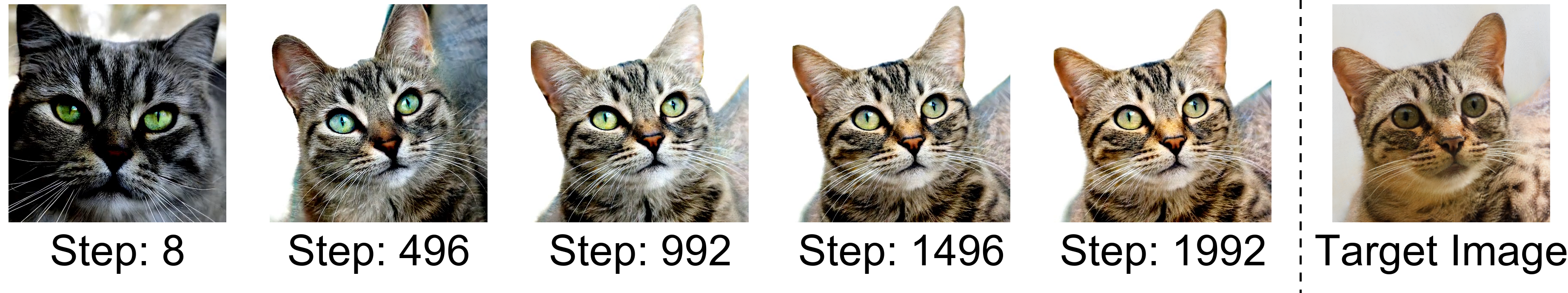}
    \\[-0.25em]
  \end{minipage}

  \begin{minipage}{0.99\linewidth}
    \centering
    \includegraphics[width=\linewidth]{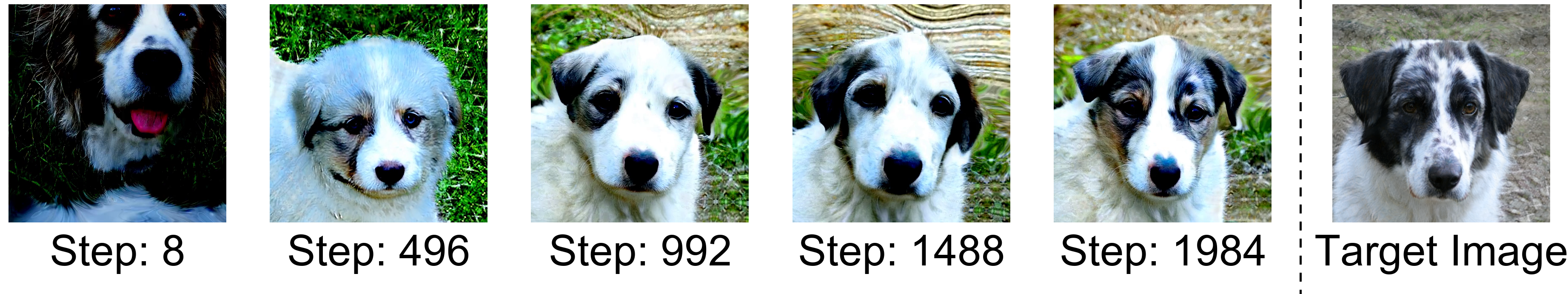}
    \\[-0.25em]
  \end{minipage}

  \caption{(Animal) Inversion attack reconstruction results of the original image from embeddings produced by a cutting-edge black-box pretrained image encoder PE~\cite{bolya2025perception}.}
  \label{fig:inversion_attack}
\end{figure}

\subsection{Results and Analysis}

% \tianyuan{Results for personal faces.} \tianyuan{Dynamic lines for personal faces.}

As shown in \cref{fig:inversion_attack_human,fig:inversion_attack}, a pretrained GAN enables successful inversion of the original input to a black-box image encoder with only the embedding of the target image as known information. This empirically supports our claim that storing non-raw data on remote servers still poses significant privacy risks.

\begin{figure}[!tbp]
  \centering
  % \begin{minipage}{0.99\linewidth}
  %   \centering
  %   \includegraphics[width=\linewidth]{figures/liangyue/plot/280_19_pe.png}
  %   \\[-0.25em]
  %   {\small (a) seed 280 to 19}
  % \end{minipage}

  % \begin{minipage}{0.99\linewidth}
  %   \centering
  %   \includegraphics[width=\linewidth]{figures/liangyue/plot/380_29_pe.png}
  %   \\[-0.25em]
  %   {\small (b) seed 380 to 29}
  % \end{minipage}
  
  \begin{minipage}{0.99\linewidth}
    \centering
    \includegraphics[width=\linewidth]{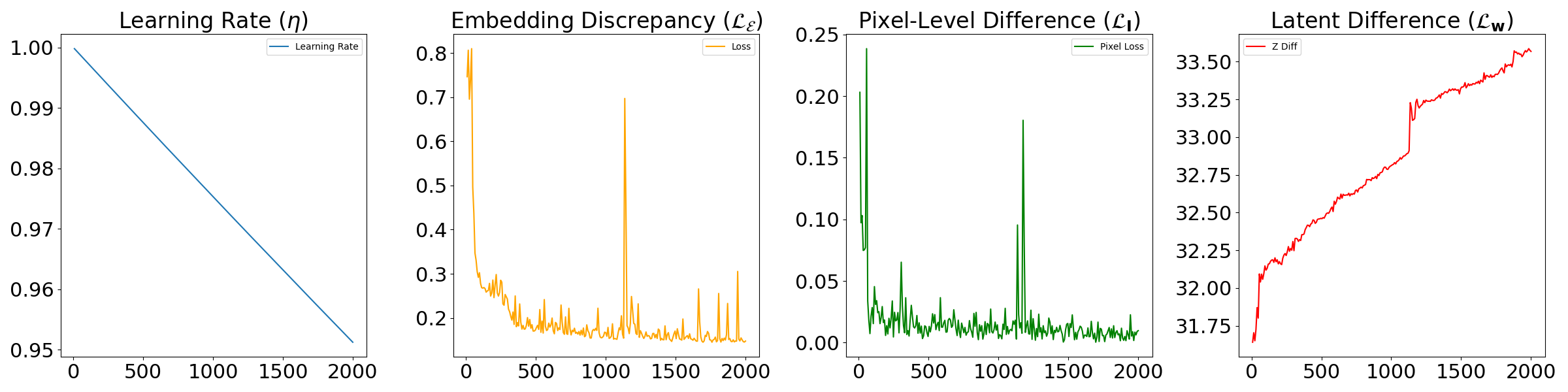}
    \\[-0.25em]
    % {\small (c) seed 280 to 18}
    {\small (a) Row $1$ in \cref{fig:inversion_attack_human}}
  \end{minipage}

  \begin{minipage}{0.99\linewidth}
    \centering
    \includegraphics[width=\linewidth]{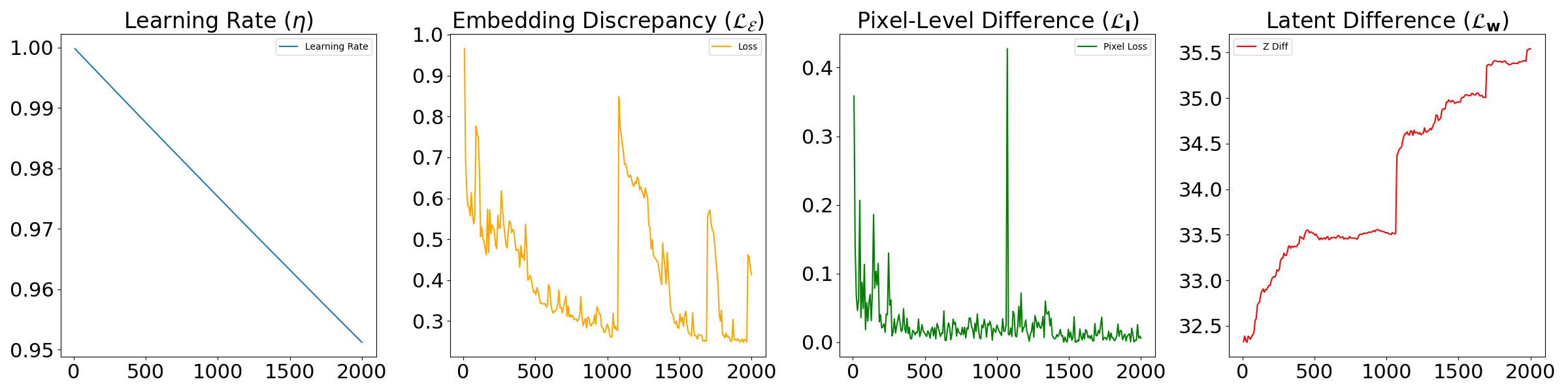}
    \\[-0.25em]
    % {\small (d) seed 300 to 22}
    {\small (b) Row $2$ in \cref{fig:inversion_attack_human}}
  \end{minipage}

  \begin{minipage}{0.99\linewidth}
    \centering
    \includegraphics[width=\linewidth]{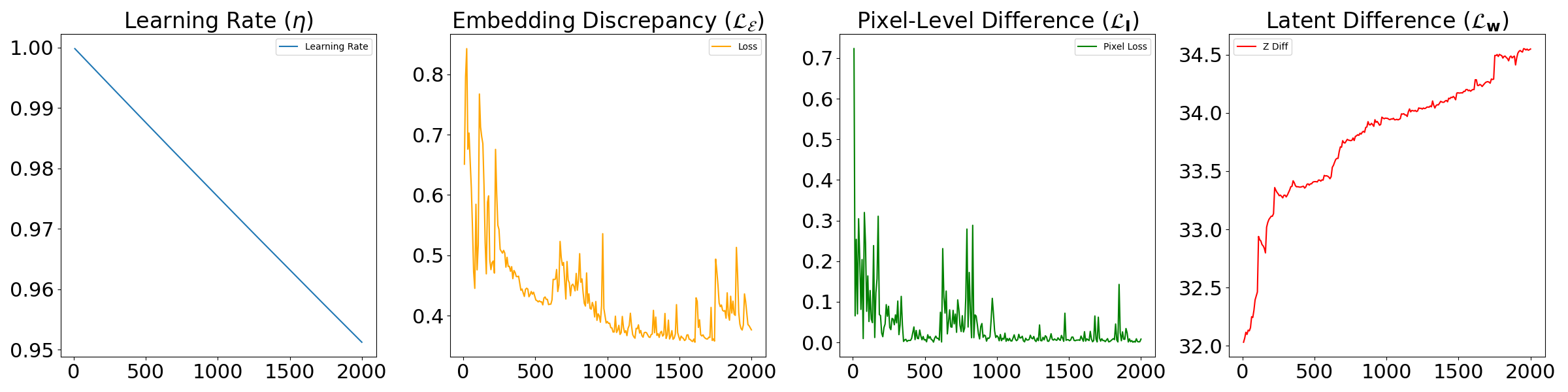}
    \\[-0.25em]
    % {\small (e) seed 500 to 28}
    {\small (c) Row $3$ in \cref{fig:inversion_attack_human}}
  \end{minipage}
  
  \begin{minipage}{0.99\linewidth}
    \centering
    \includegraphics[width=\linewidth]{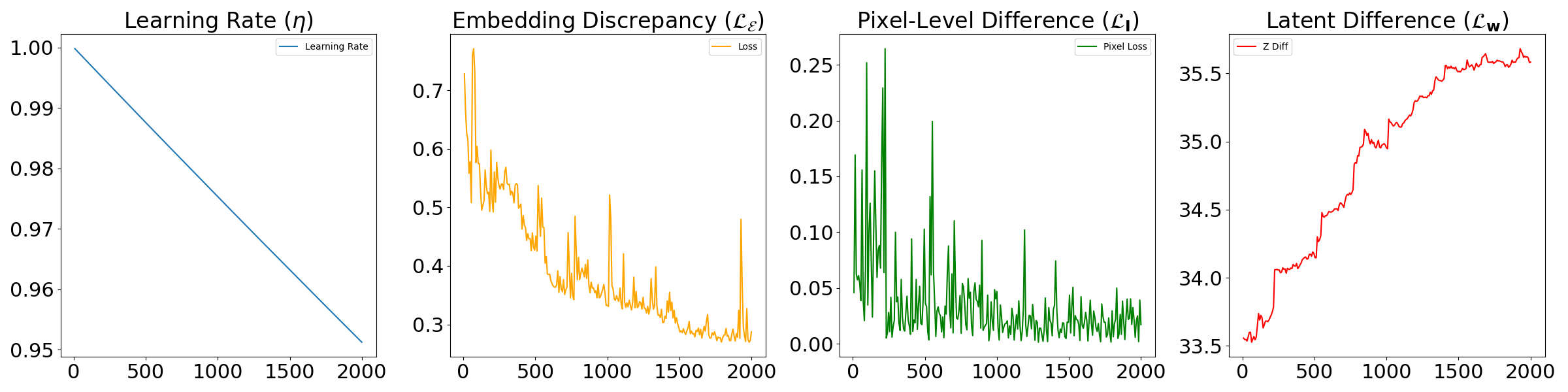}
    \\[-0.25em]
    % {\small (f) seed 280 to 25}
    {\small (d) Row $4$ in \cref{fig:inversion_attack_human}}
  \end{minipage}

  \caption{Visualization of the learning rate (left) and three metrics: embedding discrepancy $\mathcal{L}_\mathcal{E}$ (middle left), pixel-level difference $\mathcal{L}_\mathbf{I}$ (middle right) and latent difference $\mathcal{L}_\mathbf{w}$(right) during the optimization process for samples in \cref{fig:inversion_attack_human}.}
    
  \label{fig:inversion_attack_plots_human}
\end{figure}

\begin{figure}[!tbp]
  \centering
  \begin{minipage}{0.99\linewidth}
    \centering
    \includegraphics[width=\linewidth]{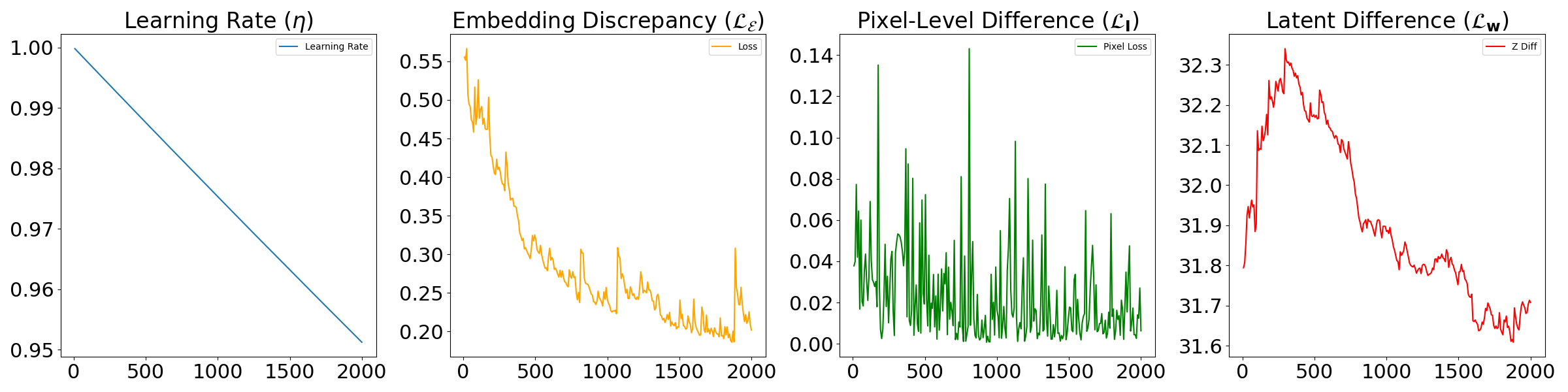}
    \\[-0.25em]
    {\small (a) Row $1$ in \cref{fig:inversion_attack}}
  \end{minipage}

  \begin{minipage}{0.99\linewidth}
    \centering
    \includegraphics[width=\linewidth]{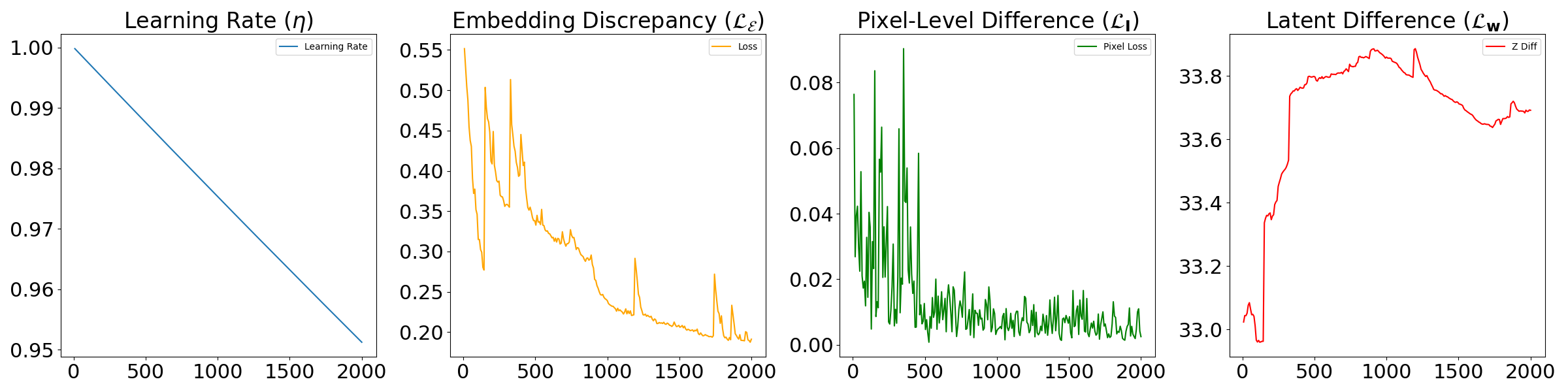}
    \\[-0.25em]
    {\small (b) Row $2$ in \cref{fig:inversion_attack}}
  \end{minipage}
  
  \begin{minipage}{0.99\linewidth}
    \centering
    \includegraphics[width=\linewidth]{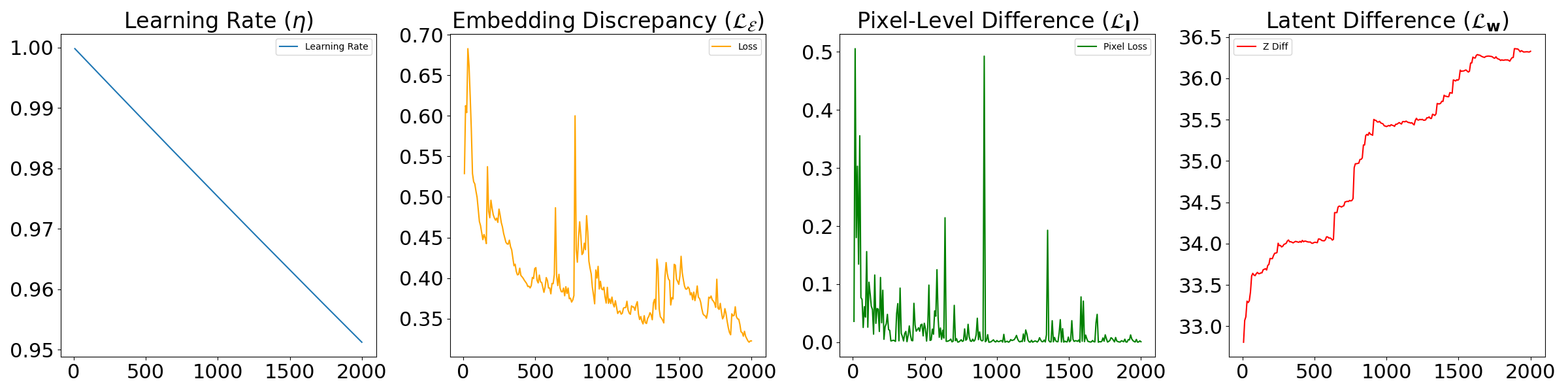}
    \\[-0.25em]
    {\small (c) Row $3$ in \cref{fig:inversion_attack}}
  \end{minipage}

  \caption{Visualization of the learning rate (left) and three metrics: embedding discrepancy $\mathcal{L}_\mathcal{E}$ (middle left), pixel-level difference $\mathcal{L}_\mathbf{I}$ (middle right) and latent difference $\mathcal{L}_\mathbf{w}$(right) during the optimization process for samples in \cref{fig:inversion_attack}.}
    
  \label{fig:inversion_attack_plots}
\end{figure}

Moreover, the optimization dynamics, including the learning rate and the three metrics ($\mathcal{L}_\mathcal{E}$, $\mathcal{L}_\mathbf{I}$, and $\mathcal{L}_\mathbf{w}$), for the samples in \cref{fig:inversion_attack_human,fig:inversion_attack} are further illustrated in \cref{fig:inversion_attack_plots_human,fig:inversion_attack_plots}. The observed fluctuations in embedding discrepancy $\mathcal{L}_\mathcal{E}$ and pixel-level difference $\mathcal{L}_\mathbf{I}$ suggest that the current method is not fully stable and is likely suboptimal.
As improving inversion performance is not our primary objective, we leave the development of more stable and optimal inversion methods to future work. 
% On the other hand, the increase trend in the latent difference $\mathcal{N}$ during the decreasing trend of $\mathcal{L}$ might be attributed to the difference in the semantic space dimension difference between that of $\mathcal{E}$ and $\mathcal{G}$.
On the other hand, the increasing trend in the latent difference $\mathcal{L}_\mathbf{w}$ alongside the decreasing trend in $\mathcal{L}_\mathcal{E}$ may be attributed to a mismatch in how $\mathcal{G}$ and $\mathcal{E}$ represent semantics in their respective embedding spaces.

% We can see that the embedding's difference $\mathcal{L}$ is not always decreasing, but it can fluctuate during the optimization process. However, the pixel-wise difference $\mathcal{D}$ keeps jumping up and down, and the noise's difference $\mathcal{N}$ even keeps increasing. This indicates that the optimization process might doesn't necessarily lead to a better reconstruction in terms of pixel-wise difference or noise's difference, even if the embedding's difference is decreasing. This also indicates that the embedding space might be more complex and non-convex, which makes the optimization process more challenging. \tianyuan{You cannot assume that similar images has similar noise for the GAN. These 2 are non-linear. Therefore, this conclusion is not convincing.}

Finally, we acknowledge that the current method may fail under poor initialization seeds, a common issue in inversion attacks~\cite{xia2022gan}. Addressing this limitation in our setting is another important direction for future work.

% \subsection{Failure Cases and Discussion}

% When the good cases are provided in Figure~\ref{fig:inversion_attack_examples}, most of the cases failed to reconstruct the original image, which is shown in \cref{fig:inversion_attack_failure}. When the image is blurry and loss of high level semantics (such as a image contains the texture of a cat but can't be recognized as a cat's face), the embedding turns to indicate something different from a cat, making it extremely difficult to reconstruct the original cat's image. 

% Furthermore, as the image generator is trained on an animal dataset, it has a strong bias towards generating images with animal features. For other types of images, such as human daily activities, it can hardly generate similar images, which makes the optimization process barely work. Therefore, a strong generator which is well trained on a large and diverse dataset is also crucial for the success of the inversion attack.

% \begin{figure}[htbp]
%   \centering

%   \begin{minipage}{0.99\linewidth}
%     \centering
%     \includegraphics[width=\linewidth]{figures/liangyue/failure/160_00_pe/summary.png}
%     \\[-0.25em]
%     {\small (a) seed 160 to 0}
%   \end{minipage}

%   \begin{minipage}{0.99\linewidth}
%     \centering
%     \includegraphics[width=\linewidth]{figures/liangyue/failure/240_11_pe/summary.png}
%     \\[-0.25em]
%     {\small (b) seed 240 to 11}
%   \end{minipage}
%   \caption{Failure cases}
%   \label{fig:inversion_attack_failure}
% \end{figure}

%%%%%%%%%%%%%%%%%%%%%%%%%%%%%%%%%%%%%%%%%%%%%%%%%%%%%%%%%%%%

\end{document}